# A semantic autonomy framework for VLM-integrated indoor mobile robots: Hybrid deterministic reasoning and cross-robot adaptive memory

Bogdan Felician Abaza *, Andrei-Alexandru Staicu, Cristian Vasile Doicin

*Faculty of Industrial Engineering and Robotics (FIIR), National University of Science and Technology POLITEHNICA Bucharest, Bucharest, Romania*

A B S T R A C T

Semantic indoor navigation requires robots to connect human instructions with spatial destinations, execute missions and retain useful operational experience. This paper presents the Semantic Autonomy Stack (SAS), a six-layer integration framework, and develops its semantic-reasoning, mission-interface and operational-memory layers over a previously validated ROS 2/Nav2 foundation. A deterministic-first cascade combines the route graph, semantic objects, robot state and stored instruction–destination associations, escalating unresolved or guarded cases to a pretrained Vision-Language Model (VLM). SAS separates destination selection, action validation, navigation execution and post-arrival confirmation. Recurrent model-mediated associations can pass a frequency-and-consistency gate and be compiled for deterministic M3 reuse; this gate measures sampled-output stability rather than semantic truth. The frozen v4.8-review implementation was evaluated on two physical robot configurations in one indoor environment, with onboard navigation and perception and shared-workstation reasoning and memory. Of 216 decision cycles outside the start gate, 112 resolved deterministically at sub-millisecond resolver-compute latency. A sealed 28-instruction challenge yielded 17 acceptable outcomes; five fast-path decisions included one incorrect selection. One promoted association was reused on the second platform through deliberate shared-digest installation. A controlled 594-observation offline comparison showed greater agreement with the stored associations and lower decision latency from M3 reuse in five familiar cases. The M3-disabled cascade reduced VLM calls but matched direct-VLM decision quality in standardized context. Large latency savings were concentrated in paths avoiding VLM inference. Together, the results establish a working integration of selective hybrid reasoning and reusable operational memory, while identifying implementation and evaluation boundaries for further development within SAS.

## 1. Introduction

### *1.1. Motivation and problem statement*

Autonomous indoor mobile robots can execute metric navigation reliably using established ROS 2 and Nav2 components [1,2], but task-level human instructions are often expressed in semantic rather than geometric terms. An operator may request a workstation, a place associated with a particular object or attribute, or a destination satisfying a contextual need without supplying a map coordinate. In such cases, navigation requires an additional layer that relates natural-language intent to the robot's graph-grounded representation of the environment. This need is increasingly relevant in industrial and human–robot settings in which multimodal language models are being explored as interfaces between human intent, spatial context, and robot action [3,4].

Vision-language models provide a flexible mechanism for interpreting visually and semantically underspecified requests, but using a large model for every navigation instruction is not always necessary or desirable. Some requests can already be resolved from explicit graph nodes, known semantic objects, lexical attributes, or previously established instruction–node associations. Other requests remain ambiguous, conflict with the available context, contain exclusions, or require richer visual-semantic interpretation. A practical semantic-navigation architecture therefore needs not only a capable deliberative model, but also a principled way to decide when deterministic evidence is sufficient and when model-based reasoning should be invoked.

A second problem is persistence. If every model-mediated decision disappears after the current run, repeated instructions continue to incur the same deliberative process and useful operational regularities cannot be reused across sessions. Conversely, converting repeated outcomes directly into deterministic rules creates its own risk: recurrence does not establish semantic correctness. Persistent memory therefore requires explicit scope, promotion criteria, safeguards, and provenance rather than an undifferentiated history buffer.

* Corresponding author.
*E-mail address:* bogdan.abaza@upb.ro (B.F. Abaza).

The problem addressed in this work is consequently the integration of three functions within one deployable robot system: deterministic-first semantic resolution with explicit abstention and escalation, constrained VLM reasoning for cases that remain unresolved or guarded, and scoped operational memory whose promoted instruction–node associations can be deliberately compiled and reused. The reported implementation preserves robot-side sensing and metric Nav2 execution while higher-layer semantic reasoning and memory services run on a shared workstation. The objective is not to replace the established navigation stack, but to extend it with a traceable higher-layer mechanism for semantic target selection, selective model use, and reusable operational knowledge.

### 1.2. Research gap

The literature reviewed in Section 2 covers all of the individual ingredients relevant to this problem, but generally emphasizes them in different combinations. VLM/VLA systems demonstrate increasingly capable embodied interpretation and action generation; selective or dual-process architectures reduce deliberative effort for easier cases; semantic and agent-memory systems retain structured spatial or episodic information; and LLM–ROS frameworks provide interfaces between foundation models and conventional robot middleware. Multi-robot systems further demonstrate shared perception, planning, or model services. These directions establish the feasibility of the component ideas, but they do not by themselves define the integration and evidence model required here.

The gap addressed by SAS is therefore not the absence of semantic navigation, memory, selective reasoning, or ROS-based model integration individually. It is the combination of these functions under an explicit decision boundary: a deterministic resolver grounded in the graph and memory first attempts only what its rules can accept;

**Table 1**
Architectural positioning of SAS against selective-reasoning, semantic-memory and LLM/VLM–ROS systems. Entries summarize published capabilities and validation scope, not benchmark-equivalent performance; external systems were not rerun. Not documented means that the source does not establish the capability; partial denotes limited coverage. SAS model substitution supports L3b interface portability, not semantic model invariance.

(a) System architecture, reasoning, and model dependence

| System | System / task scope | Fast-path / selection mechanism | LLM/VLM role, abstention and safeguards | Task-specific model adaptation | Model portability evidence |
|---|---|---|---|---|---|
| IROS dual-process system [6] | Indoor navigation on-device; 5 real buildings | Key-frame gate; VLM only when match is ambiguous | VLM invoked on ambiguity; safety maneuver module | Pretrained VLM; training not documented | Two VLMs evaluated under matched images / prompts |
| Hydra-Nav [7] | Object-goal navigation in simulation; real-robot demo | One VLM switches mode on its own token | Internal VLM reasoning; safeguards not documented | Three-stage curriculum fine-tuning of a 7B VLM | not documented |
| CausalNav [8] | Outdoor navigation; simulation and campus trials | Hierarchical planner; selection not documented | LLM builds the graph; safeguards not documented | Task-specific model training not documented | Five LLM backends evaluated in simulation |
| EchoVLA [9] | Mobile manipulation; simulation and one real arena | Single per-part diffusion policy, memory conditioned | Abstention and safeguards not documented | Diffusion policy trained on a purpose-built corpus | not documented |
| ROSClaw [10] | General ROS 2 task execution; three morphologies | No fast path; every command reaches the agent | Validator allows or blocks before execution | Pretrained models behind one provider interface | Four backends evaluated under identical prompts |
| Prior work by the authors [5] | Indoor route planning on CPU-only edge; 3 platforms | Route choice by edge penalty; no resolver cascade | No LLM/VLM stage in the architecture | No learned parameters in ASF; detector not documented | not documented |
| This work - SAS | Indoor semantic target selection with ROS 2 / Nav2 execution; two physical robot configurations, one indoor site; frozen v4.8-review campaign | Seven-step deterministic resolver with M3 Step 0; no separately trained routing classifier | Selective L3b escalation after deterministic abstention; negation / exclusion precedence, location-conflict handling, Step-0 threshold and margin | Pretrained VLM; no task-specific fine-tuning or separately trained task classifier | Frozen L3b interface replayed on 3 VLMs × 3 seed bases; interface portability demonstrated, semantic invariance not claimed |

(b) Memory lifecycle and reuse

| System | Memory representation and scope | Runtime memory use | Promotion / consolidation | Persistence and shared reuse |
|---|---|---|---|---|
| IROS dual-process system [6] | Rule table of conditions and actions, per trip | Table matched by cosine similarity to pick an action | Per-trip table generation; promotion not documented | Per-trip only; persistence not documented |
| Hydra-Nav [7] | Landmark graph of image-text nodes, within episode | Memory is input to the planning mode each cycle | Nodes appended and pruned; promotion not documented | Within episode; cross-session persistence not documented |
| CausalNav [8] | Multi-level embodied graph built by an LLM | RAG retrieval over the graph for open queries | Time-window graph update; promotion not documented | Graph per environment; sharing not documented |
| EchoVLA [9] | Scene voxel memory plus episodic token buffer | Coarse-to-fine attention over both memory banks | Update gated by a reconstruction threshold | Scene memory refined across episodes, per site |
| ROSClaw [10] | Runtime memory store, reset for each trial | Sliding context window; memory unused across trials | not documented | Memory reset each trial; shared reuse not documented |
| Prior work by the authors [5] | Object-level GeoJSON map with mobility classes | Map annotates edges with penalties and speed limits | EMA position fusion and TTL expiry; no promotion gate | Persists across restarts; not shared between robots |
| This work - SAS | Typed M1–M5 operational memory; M3 learned instruction-node associations | M1/M2 contextual prompt content + M3 deterministic Step-0 matching | M3 gate: frequency / consistency / accepted L3b | Compiled persistent memory; demonstrated shared-memory reuse across robot platforms for one induced association |

Note. Validation coverage is asymmetric across memory classes: M3 is directly evaluated; M1 and M2 are ablated only jointly with M3; M4 and M5 are not independently evaluated.

unresolved, negated, or conflicting cases are escalated through a constrained VLM application interface; accepted model-mediated outcomes can later enter a scoped promotion workflow; and promoted M3 associations can be compiled, installed, and reused deterministically across compatible robot platforms. The promotion gate is intentionally separated from semantic correctness, and cross-platform reuse occurs through a managed memory artifact rather than autonomous peer-to-peer learning.

A second gap concerns validation scope and reproducibility. A layered architecture can easily blur what is inherited, what is newly introduced, and which mechanisms have actually been tested. In the present work, the lower navigation/perception foundation is inherited from the authors' prior experimentally validated system [5], while the revised contribution concentrates on the higher semantic-reasoning, mission-interface, and operational-memory layers. The final-system evidence is likewise tied to a frozen v4.8-review implementation with recorded software, model, graph, ontology, and parameter identity. This separation allows the claims to be stated at the level supported by the experiments rather than treating end-to-end operation as independent validation of every layer or memory component.

Accordingly, the present study investigates whether a deterministic-first semantic resolver, explicit safeguard and abstention logic, scoped operational memory, and shared-memory reuse across robot platforms can be integrated with an inherited ROS 2/Nav2 navigation foundation and evaluated as a frozen higher-layer system on physical robots. Table 1 positions this integration against representative related systems; Sections 3–7 then separate architecture, mechanisms, experimental design, and evidence.

### *1.3. Contributions*

Within the scope defined above, this paper makes four contributions.

1. Six-layer SAS integration framework. We define the Semantic Autonomy Stack as a capability and integration framework spanning the mobile platform (L0), navigation execution (L1), perception and semantic map (L2), semantic reasoning (L3), mission interface (L4), and operational memory (L5). Its layer boundaries connect human instructions, graph-grounded destinations, robot execution and accumulated experience through explicit interfaces. This work develops and evaluates L3–L5 over the previously validated L0–L2 foundation [5], providing a framework in which reasoning and memory mechanisms can evolve while retaining their navigation and perception interfaces.
2. Hybrid deterministic–VLM semantic reasoning. We introduce a seven-step deterministic-first resolver combining graph, semantic-object, robot-state and promoted-association context with constrained VLM escalation. The integration includes threshold-and-margin defer decisions, negation/exclusion precedence, location-conflict handling and post-response action validation. Destination selection, navigation completion and post-arrival semantic/visual confirmation remain distinct events. A controlled comparison with direct VLM resolution and an M3-disabled hybrid configuration evaluates both decision quality and selective invocation, separating resolver computation from offline instruction-to-decision latency. The pipeline uses pretrained VLMs without task-specific fine-tuning or a separately trained task classifier.
3. Operational-memory organization with M3 promotion and reuse. We define the M1–M5 organization and an audit-to-memory lifecycle in which recurrent model-mediated instruction–node outcomes become candidate M3 associations, pass an explicit frequency/consistency/model-path gate, and are compiled into a runtime digest for subsequent deterministic reuse. In this work, “adaptive” refers to the promotion and compilation of recurrent model-mediated instruction–node associations into deterministic rules for subsequent reuse, rather than to online model learning or validated acquisition of human preferences. This adaptation changes the stored associations available to the resolver without retraining the VLM. Promotion establishes sampled-output stability; semantic validation is a separate requirement. We demonstrate reuse of one induced association on a recipient robot through deliberate shared-digest installation. Cross-platform use is therefore managed shared-memory reuse, rather than autonomous robot-to-robot learning. The controlled comparison assesses M3′s contribution separately from retained M1/M2 context.
4. Physical integration and evidence across complementary evaluations. We implement the higher-layer framework in the frozen v4.8-review system on two physical robot configurations in one indoor environment. The confirmatory Session E campaign evaluates selective reasoning, guard behavior, confirmation and shared-memory reuse, separately from earlier development-stage observations. Controlled offline comparisons, independent destination adjudication and targeted matcher/guard diagnostics supplement the physical evidence, with outcome definitions, experimental units and reproducibility boundaries stated for each evaluation. These studies assess the implemented mechanisms and identify concrete development priorities within the framework.

The contributions distinguish the SAS architecture, the frozen implementation and the experimental coverage. M3 is the directly demonstrated adaptive-memory mechanism; M1/M2 are retained as context in the controlled M3 comparison, while their individual performance contributions and those of M4/M5 are not isolated. The model-substitution replay assesses portability of the L3b application interface across the tested VLMs. Environmental generalization and fleet-scale operation remain outside the evaluated coverage.

### *1.4. Paper organization*

Section 2 reviews related work and positions SAS. Section 3 defines the six-layer framework and separates the inherited L0–L2 foundation from the higher-layer contribution. Sections 4 and 5 describe hybrid semantic reasoning and scoped memory. Section 6 defines the versioned evidence model, experimental design and analysis, including the external localization reference, model-substitution replay, controlled semantic comparison and matcher/guard diagnostics. Section 7 reports their results, retaining Sessions A–C separately as development history. Section 8 discusses interpretation and limitations; Section 9 concludes.

## 2. Related work

Related work is organized around VLM-based semantic navigation, selective reasoning, persistent semantic/agent memory and LLM/VLM–ROS integration. The comparison examines how systems assign reasoning, memory, validation and execution responsibilities; heterogeneous protocols preclude performance rankings.

### *2.1. VLM-based semantic and embodied navigation*

Vision-language models have increasingly been used to bridge the gap between geometric navigation and task-level instructions expressed in natural language. Recent semantic-navigation and VLA surveys [4,11,12] distinguish systems in which a multimodal model participates directly in action generation from modular architectures in which model reasoning augments an otherwise conventional perception–-planning–control stack. These two directions expose a common design question for mobile robots: how much semantic interpretation should remain inside a large multimodal model, and how much can be delegated to explicit spatial representations and conventional navigation components.

End-to-end vision-language-action approaches integrate visual observations, language, and action prediction within a learned policy.

Examples include action-language formulations for adapting VLMs to robot control [13] and memory-augmented VLA architectures such as MemoryVLA [14]. These systems demonstrate the value of large multimodal models for embodied control and long-horizon behavior, particularly in manipulation-oriented settings. Their architectural emphasis, however, is different from a graph-grounded indoor navigation system in which target resolution can be separated from metric route execution and in which a large model need not participate in every instruction cycle.

Modular navigation systems retain this separation more explicitly. Chandaka et al. [15] use VLM reasoning for human-like navigation behaviors in environments designed for people, while VLM-Social-Nav [16] introduces model-derived social-compliance information into a navigation pipeline. Wang et al. [17] combine multimodal language reasoning and spatial scene understanding for human-guided mobile-robot navigation in smart-manufacturing settings. In these examples, VLM reasoning augments navigation with capabilities that are difficult to express through geometric goals alone, such as interpreting landmarks, human-centered scene semantics, or contextual constraints.

Multi-robot work further broadens the role of multimodal models. Co-NavGPT [18], for example, uses shared semantic information and VLM-based planning for cooperative visual semantic navigation. This addresses collaboration and exploration across robots, whereas SAS considers a different higher-layer problem: repeated semantic target-resolution knowledge can persist across sessions and can be reused through a shared compiled memory artifact by another robot operating against compatible graph and semantic identifiers. The latter mechanism is described conservatively as shared-memory reuse rather than autonomous robot-to-robot learning.

Evaluation methodology remains heterogeneous across this literature. NaviTrace [19] and retrieval-augmented navigation approaches such as RANa [20] illustrate the importance of separating semantic goal interpretation, memory retrieval, and path generation when evaluating model-based navigation. Consequently, the present manuscript does not treat reported results from simulation benchmarks, real-building navigation, manipulation environments, and the SAS indoor campaign as directly comparable performance measurements. Their value here is architectural: they establish the range of ways in which multimodal reasoning can be inserted between human intent and robot action.

### 2.2. Selective reasoning and dual-process architectures

Selective reasoning aims to reserve expensive deliberative processing for cases that cannot be handled by a cheaper or more constrained mechanism. The general motivation is consistent with dual-process accounts of fast and deliberative reasoning [21], but robotic implementations differ substantially in how the routing decision is made and in what constitutes the fast path.

The IROS dual-process indoor-navigation system [6] is the closest comparator in this respect. Its fast pathway uses key-frame comparison and condition matching against a goal-specific Condition-to-Action table generated during initialization; zero or multiple applicable actions trigger the slower VLM pathway. The important comparison with SAS is therefore not a reported fast-path percentage, because the instruction distributions and protocols differ, but the routing mechanism itself. The IROS system uses condition matching over a VLM-generated action table, whereas SAS uses a fixed seven-step deterministic resolver grounded in the navigation graph, semantic objects, and promoted M3 instruction–node associations.

Hydra-Nav [7] places the fast/slow distinction inside a single VLM-based navigation policy. Its fast mode performs direct action generation, while an observation token can trigger slower chain-of-thought reasoning supported by landmark and short-term memory. The documented training pipeline includes task-specific alignment, supervised fine-tuning, and iterative rejection fine-tuning. This contrasts with the reported SAS reasoning pipeline, which uses pretrained VLM inference only after the deterministic cascade abstains or a safeguard requires escalation and introduces no task-specific VLM fine-tuning or separately trained task classifier. A recent semantically gated visual-servoing approach applies a related gating principle to lightweight VLM navigation, but gates continuous visual control toward an identified target rather than discrete instruction-to-graph-node resolution [22].

Other systems adapt reasoning effort rather than providing a separate deterministic semantic resolver. CogDDN [23] varies reasoning depth according to cognitive demand, and dual-process VLA work [24] applies related fast/slow concepts to manipulation. These approaches reinforce the broader motivation for selective computation, but the location of the selection mechanism differs: it may reside in a learned model, in a learned switching policy, in a calibrated uncertainty measure that triggers deferral [25], or, as in SAS, in an explicit resolver whose acceptance, abstention, and safeguard conditions are inspectable.

This distinction also affects what can be claimed from evaluation. A deterministic-before-VLM architecture should not be judged by comparing naturally selected fast-path and model-path cases as if they were matched experimental arms, and cross-paper latency or success values from different protocols do not form a controlled benchmark. The revised manuscript therefore uses the structured comparison in Table 1 to compare mechanisms and validation scope, while the performance of the frozen SAS resolver is evaluated only within its own declared experimental design.

### 2.3. Semantic memory and agent memory

Memory in embodied systems spans several functions that should not be conflated: persistent environment representation, retrieval of prior observations, episodic history, online graph updating, and the conversion of repeated outcomes into reusable decision structures. Recent agent-memory surveys [26] emphasize similarly broad distinctions among memory representation, retrieval, update, and consolidation. ExpeL similarly distills accumulated agent experience into reusable rules without updating model parameters [27]. Recent work on experience reuse further reports that abstract procedural memories can transfer more effectively than detailed trajectories, supporting the use of compact reusable memory representations rather than raw interaction histories [28]. For mobile robots, the relevant question is therefore not simply whether a system “has memory,” but how that memory is scoped, how it is updated, and whether it changes runtime decision making.

Structured semantic maps are one important form of persistent memory. MapNav [29] replaces raw historical observations with annotated semantic maps that can be retrieved during VLM-based vision-and-language navigation. Meta-Memory [30] similarly addresses semantic-spatial retrieval by organizing prior observations so that a robot can recover where relevant entities were previously encountered. Such methods use structured memory primarily to improve later reasoning or spatial recall.

CausalNav [8] provides a more dynamic example. Its multi-level Embodied Graph represents buildings, objects, and trajectories; the graph is updated online and queried by hierarchical LLM reasoning to support target and route selection. The primary-source audit performed for the revised comparison distinguishes this online graph maintenance from demonstrated cross-session shared memory: historical state is used during operation, but persistent reuse across independent sessions or across robots is not established by the cited study.

EchoVLA [9] combines a persistent environment-specific scene voxel memory with an episodic token buffer for long-horizon mobile manipulation. Retrieved memory conditions a trained diffusion policy, while memory updates are gated by a reconstruction threshold. This is a strong example of memory as part of a learned action-generation process. SAS uses memory differently: its M1–M5 organization separates operational information by role and scope, and M3 can move selected repeated model-mediated instruction–node outcomes into a compiled deterministic reuse path. Promotion is therefore an explicit lifecycle operation rather than merely retrieval of a larger context store.

The revised SAS terminology is deliberately narrower than the submitted manuscript. A promoted M3 item is a learned instruction–node association, not evidence of a validated operator preference. Likewise, cross-platform use is shared-memory reuse through deliberate digest compilation and installation, not direct robot-to-robot learning. This distinction is central to the comparison because it separates memory representation from the evidentiary question of what has actually been learned and validated. Table 1 therefore compares representation, runtime use, promotion or consolidation, persistence, and shared reuse as separate dimensions rather than treating “memory” as a single capability.

### 2.4. LLM/VLM–ROS integration and multi-robot/shared-service architectures

A second line of work addresses the interface between foundation models and robot middleware rather than semantic navigation alone. These frameworks expose ROS observations and actions to an agent, translate language into robot operations, and increasingly add validation, logging, or persistent context so that model reasoning can be integrated with conventional robot controllers.

ROSClaw [10] is the closest comparator at the executive-interface level. It integrates an agent runtime with ROS 2 and formalizes an executive contract around affordances, observations, action validation, and audit logging. The primary-source audit for Table 1 also confirms configuration-level substitution across multiple foundation-model backends under the same executive interface. SAS adopts the same ⟨A, O,V,L⟩ contract perspective but specializes it for graph-grounded semantic navigation, where the deliberative component proposes a destination and downstream validation and Nav2 execution remain explicit. The resulting portability claim is correspondingly narrow: the revised SAS evidence supports portability at the L3b application interface across the evaluated models, not semantic model invariance.

Other general-purpose language-to-ROS systems operate at a higher task-programming level. The framework reported by Mower et al. [31] translates language into embodied robot behavior with feedback and structured execution, while ROSA [32] provides a language-oriented interface for ROS introspection and mission operations and RAI [33] provides a broader agent framework for embodied AI. A recent review of real-world LLM-enabled robotic systems further characterizes such systems along dimensions including autonomy, goal-directedness, adaptability, and decision-making [34]. These systems illustrate the value of keeping language-model reasoning above stable middleware interfaces rather than embedding it inside low-level control.

Multi-robot integration introduces an additional systems concern: whether shared model or planning services can be used by more than one physical platform without conflating shared reasoning infrastructure with shared learned state. Cooperative navigation systems such as Co-NavGPT [18] share observations and semantic state to coordinate exploration. In SAS, the two forms of sharing are separated architecturally: the L3b inference service can be shared by the robot configurations, while persistent cross-platform memory reuse occurs through a compiled memory artifact with explicit installation and platform-aware attribution. The revised manuscript therefore treats concurrent shared-service operation and shared-memory reuse as distinct capabilities.

This separation also limits the generality of the deployment claim. The cited agentic ROS frameworks cover broader robot morphologies and task classes, whereas SAS is domain-specialized to indoor semantic target resolution on top of ROS 2/Nav2. Conversely, the SAS design couples the executive interface to a deterministic semantic resolver and an explicit M3 promotion/reuse lifecycle. The relevant novelty question is therefore the integration of these mechanisms and their validation scope, not whether SAS is a more general agent framework than the systems above.

### 2.5. Positioning of the present work

Table 1 consolidates the comparison along the dimensions most relevant to the revised contribution: system/task scope, fast-path or selection mechanism, the role of the LLM/VLM and safeguards, memory representation and runtime use, promotion or consolidation, persistence and shared reuse, task-specific model adaptation, and model-portability evidence. The table is intentionally organized by capability and scope. The external systems were not rerun in this study, and their environments, task sets, models, instruction distributions, and experimental protocols are not benchmark-equivalent.

The comparison shows that the individual ingredients of SAS have clear precedents but are usually emphasized separately. The IROS system [6] and Hydra-Nav [7] address selective fast/slow reasoning; CausalNav [8] and EchoVLA [9] provide structured, dynamically used embodied memory; ROSClaw [10] provides a reusable executive/validation interface and controlled backend substitution; and the authors' prior work [5] provides the inherited physical navigation, semantic mapping, and semantic-aware routing foundation used by the present system. The revised contribution is therefore positioned as an integration of a deterministic-first semantic resolver, explicit safeguard and abstention behavior, scoped operational memory with M3 promotion, and shared-memory reuse across robot platforms on top of that inherited ROS 2/Nav2 foundation.

The comparison does not establish benchmark superiority or infer absent capabilities from undocumented features. Model substitution supports L3b application-interface portability rather than equivalent semantics. Section 3.4 maps each contribution to its evidence, distinguishing architectural scope from independently evaluated layers and memory components.

## 3. Semantic autonomy stack

The Semantic Autonomy Stack (SAS) is a six-layer integration framework defined by responsibilities and interfaces rather than one prescribed implementation. L0–L2 provide the inherited navigation, perception and semantic-map foundation [5]; this study develops and evaluates the higher-layer reasoning, mission interface and operational memory. The following sections define the layer interfaces, relate them to the foundation, instantiate v4.8-review and map the contributions to their evidence.

### 3.1. SAS layer model and interfaces

The SAS organizes the capabilities required for semantically autonomous mobile-robot navigation into six layers (Fig. 1), ordered from the physical platform (L0) to operational memory (L5). The layered organization separates metric navigation and perception from semantic intent resolution, external mission entry, and persistent operational knowledge. In the present work, L0–L2 are treated as inherited infrastructure, while L3–L5 define the higher-layer integration focus.

L0 - Mobile platform. L0 comprises locomotion hardware, onboard sensing, power, low-level control, and the compute resources physically associated with the robot. The SAS does not prescribe a particular mobile morphology; rather, L0 provides the platform-specific capabilities on which the higher layers depend.

L1 - Navigation execution. L1 provides localization, route or path planning, path following, and local obstacle avoidance. It receives an admissible navigation goal or route request from the semantic layer and executes that request in metric space. In the reported implementation, L1 is realized with ROS 2/Nav2 and operates on the robot independently of the higher-layer semantic interpretation of the requested destination.

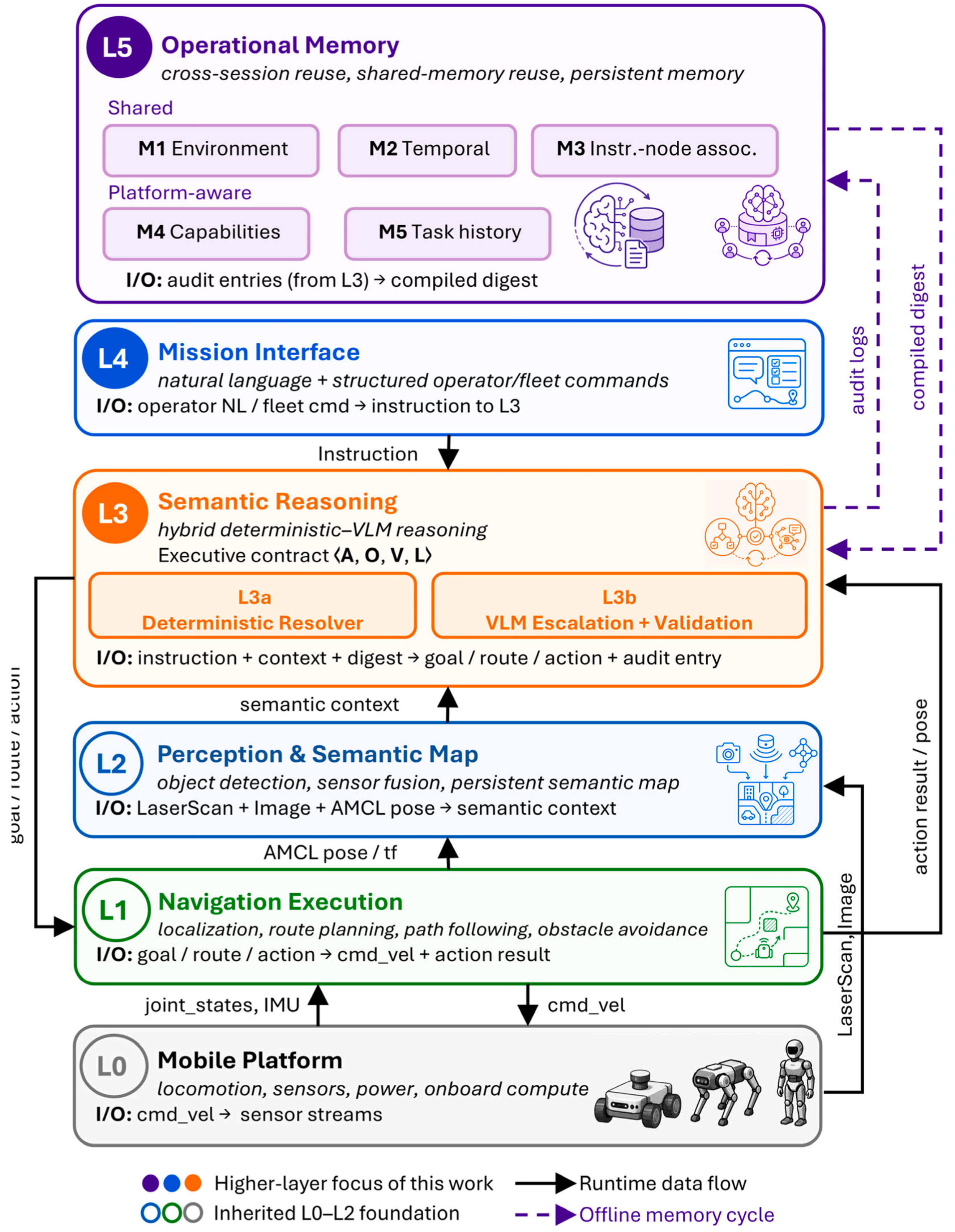

**Fig. 1.** The Semantic Autonomy Stack. Hollow badges identify the inherited L0–L2 foundation [5]; filled badges identify the L3–L5 focus of this work. Solid arrows show runtime data flow; dashed arrows show the offline audit-to-digest cycle supporting subsequent M3 Step-0 reuse. The executive contract ⟨A, O, V, L⟩ follows ROSClaw [10] (Section 4.1). Shared and Platform-aware are architectural summaries; exact M1–M5 scopes appear in Section 5. The framework accommodates mobile platforms generally; physical validation here covers two differential-drive configurations in one indoor environment (Section 6).

L2 - Perception and semantic map. L2 transforms sensor observations into localized semantic context that can be consumed by L3 and, where appropriate, can also influence L1 route selection. A concrete L2 instance may combine object detection, geometric sensor fusion, and a persistent semantic map. The implementation inherited in this study uses the monocular-camera–2D-LiDAR Angular Sector Fusion (ASF) pipeline, persistent semantic mapping, and semantic annotation of the Nav2 Route Server graph established in our prior work [5].

L3 - Semantic reasoning. L3 bridges operator intent and executable robot action. It receives a natural-language instruction together with graph, semantic-object, robot-state, and memory context and resolves the instruction to an admissible target or to an explicit abstention/clarification outcome. The present implementation divides L3 into a deterministic-first resolver (L3a) and a selectively invoked VLM escalation-and-validation path (L3b). Post-arrival semantic/visual confirmation is kept distinct from the initial target-selection decision. The present implementation adopts the ROSClaw executive-contract formalization ⟨A, O, V, L⟩ [10] and specializes it for semantic navigation; its components and their role in the v4.8-review reasoning pipeline are defined in Section 4.1. The detailed v4.8-review resolver, safeguards, escalation path, and confirmation logic are defined in Section 4.

L4 - Mission interface. L4 is the boundary between external operators or supervisory systems and the semantic-reasoning pipeline. It accepts mission requests, including natural-language and structured operator/fleet commands, and forwards them through a stable internal interface to L3. This separates mission entry from the robot-specific navigation implementation and allows the same higher-layer logic to be used with either robot configuration considered here. At the framework level, L4 is designed to accommodate standardized fleet-management interfaces, including VDA 5050 [35].

L5 - Operational memory. L5 provides the operational-memory functions of SAS through the M1–M5 scoped semantic-memory framework. It maintains scoped persistent operational information supporting cross-session reuse and, where explicitly supported, shared-memory reuse across robot platforms, and it includes audit-derived learned instruction–node associations, compiled memory used by L3, and operational logging/analysis functions. The compact Shared and Platform-aware group labels in Fig. 1 summarize the deployment-level organization; the exact scope, lifecycle, and asymmetric experimental coverage of each memory layer are defined in Section 5.

Fig. 1 summarizes both the six-layer reference structure and the deployment interfaces. Hollow layer badges identify the inherited L0–L2 foundation, whereas filled badges identify the higher-layer focus of this work. Runtime information flows upward from sensing, localization, the route graph, and semantic objects to the reasoning layer, while admissible goals and actions flow back to onboard navigation. In parallel, audit records support an offline memory cycle whose compiled output can be loaded by subsequent higher-layer runs. The layer boundaries are intended to limit coupling and make implementation choices explicit; they do not imply that substituting a component or model within one layer preserves equivalent semantic behavior.

### 3.2. Relationship to the previously validated L0–L2 foundation

The lower part of the SAS is not introduced here as a new navigation or perception stack. Our prior Sensors study [5] developed and experimentally evaluated the Xplorer L0–L2 foundation, including the custom ROS 2 mobile platforms, Nav2-based metric navigation, monocular camera–2D LiDAR ASF object localization, a persistent semantic map, semantic graph annotation, and penalty-weighted Nav2 Route Server replanning. That study evaluated the physical perception-and-routing pipeline over 115 navigation legs on Xplorer-A, Xplorer-B, and Xplorer-C and therefore provides the empirical basis for treating these capabilities as inherited infrastructure in the present manuscript.

The revised SAS manuscript reuses the corresponding functional interfaces: L2 supplies graph-grounded semantic context, L1 executes metric navigation, and the higher layers consume those capabilities rather than redefining them. The current Xplorer-B and Xplorer-C deployment nevertheless differs in hardware and compute partitioning, and those deployment-specific differences are reported explicitly in Section 6. Such differences are treated as integration adaptations of the inherited foundation, not as new L0–L2 algorithmic contributions.

Accordingly, the experimental claims of this paper are concentrated on the higher-layer mechanisms and their integration with the inherited stack. Session E exercises the frozen system end to end, but it is not interpreted as an independent layer-by-layer revalidation of L0–L2. This distinction is carried through the contribution-to-evidence map in Section 3.4 and the experimental design in Section 6.

### 3.3. Frozen v4.8-review deployment instance and compute topology

The revised experiments instantiate SAS on two physical configurations, Xplorer-B and Xplorer-C. Both robots execute L0–L2 onboard and expose the same higher-layer semantic/navigation interfaces. Xplorer-B uses a single Raspberry Pi 5 for navigation and object detection, whereas Xplorer-C separates the detector onto a second Ethernet-connected Raspberry Pi 5 while retaining the navigation stack on the primary board. Both use the same Logitech C920 camera and STL-19P 2D LiDAR sensing modalities and are exposed to the same canonical route graph and semantic environment representation. The remaining platform differences, including IMU availability, drivetrain, motor-control hardware, and battery instrumentation, are reported in Section 6.1 rather than repeated here.

Each robot provides a per-robot context service through which the higher layers obtain the current pose, camera frame when required, route graph, and semantic-object context. Metric navigation remains robot-side: Nav2 is launched on each platform, and accepted higher-layer targets are executed through ROS 2 navigation action interfaces. Thus, the physical motion and localization loop is not performed on the AI workstation.

L3a deterministic reasoning, L3b orchestration and VLM inference, and L5 memory/audit services execute on the shared workstation. The workstation used in the reported campaigns is equipped with an Intel Core Ultra 7 255H CPU, an NVIDIA RTX 5050 GPU, 32 GB RAM, and Ubuntu 24.04. GPU acceleration is used for L3b VLM inference; the deterministic L3a path does not require GPU execution. This placement makes the deployment boundary explicit and avoids treating the full SAS inference stack as an onboard CPU-only implementation.

At runtime, L3a combines the incoming instruction with robot/context data and the installed compiled memory. When deterministic resolution escalates, L3b uses the current image together with the available semantic and robot context and invokes the configured pretrained VLM through the frozen application interface. An admissible goal, route, or action is then passed to the onboard navigation stack, while execution and confirmation outcomes are returned to the higher-layer audit trail. In the complementary offline path, eligible audit history can be processed into a compiled memory artifact and installed for later deterministic or prompt-context reuse, as detailed in Section 5.

Session E uses the frozen v4.8-review implementation. The control VLM is the pretrained qwen3.5:4b model; the reported SAS pipeline introduces no task-specific model fine-tuning or separately trained task classifier. Exact software, model, provenance, and resolver identities are consolidated in Section 6.3, and the campaign/version boundary is defined in Section 6.

### 3.4. Contribution and evidence scope

Because SAS combines previously validated lower-layer infrastructure with newly introduced higher-layer mechanisms, the contribution boundary must be stated independently of the architectural layer

**Table 2**
Contribution-to-evidence map. The inherited L0–L2 basis is separated from the higher-layer elements introduced here, together with their frozen evidence and claim scope. Experimental design is given in Section 6 and outcomes in Section 7.

| Contribution | Inherited basis | New in this work | Frozen evidence | Validation scope and limitation |
|---|---|---|---|---|
| C1 - SAS reference and integration framework | Prior Sensors work: L0–L2 physical/navigation stack, ASF perception, persistent semantic mapping, Nav2 Route Server, and penalty-weighted replanning. | L0–L5 capability framework integrating inherited L0–L2 with semantic reasoning (L3), mission interface (L4), and operational memory (L5). | Prior L0–L2 validation; Session E end-to-end operation of frozen v4.8-review. | Framework/integration validation. Experiments target higher layers; inherited L0–L2 are not newly developed or independently revalidated here. |
| C2 - Hybrid deterministic–VLM reasoning, safeguards, and confirmation | Semantic map, graph, and route execution are inherited; natural-language resolution and deterministic-before-VLM orchestration are not. | Seven-step L3a resolver with Step-0 reuse, ambiguity handling, negation/location-conflict safeguards, constrained L3b escalation, and post-arrival confirmation. | Session E safeguard, repeatability, sealed-instruction and confirmation tests; independent sealed-instruction adjudication; controlled A/B/C comparison and multi-VLM replay; historical matcher sensitivity, labeled matcher-boundary diagnostics and deployed-parser/guard branch tests. | Direct higher-layer evaluation. Deterministic resolver computation is sub-millisecond in Session E; invocation coverage and offline latency benefits depend on workload. Safeguards and the parser have documented failure boundaries. Model replay supports L3b interface portability, not semantic invariance. |
| C3 - Scoped semantic memory: M3 promotion and shared-memory reuse | Persistent semantic environment representation is inherited; the operational-memory taxonomy and compiled instruction-node reuse workflow are not. | M1–M5 scoped operational memory; M3 promotion of repeated instruction-node outcomes; deterministic Step 0; shared compiled-memory reuse across robot platforms. | Session E promotion/rejection, cross-platform installed-memory reuse and compiled-memory ablation; controlled A/B comparison with M1/M2 context retained; post-hoc promotion sensitivity and multi-VLM replay. | M3 promotion/reuse is directly tested; A–B attributes reuse benefits on five familiar items with M1/M2 fixed. The compiled-memory removal experiment jointly removes M1/M2 and M3; separate M1/M2 and M4/M5 contributions remain unevaluated. Promotion measures sampled-output stability, not semantic correctness, and can depend on model/sampling. Cross-platform reuse uses deliberate digest installation. |
| C4 - Physical validation of frozen higher-layer integration | Two physical platforms and the previously validated L0–L2 execution stack; hardware is not a new algorithmic contribution. | Higher reasoning and memory integrated on two robot configurations sharing workstation-side L3a/L3b/L5 services, with an external localization reference. | Two-platform Session E integration, shared L3b access, memory interventions, and separate external-localization characterization. | Integration validation, not standalone algorithmic novelty. Two robot configurations, one indoor environment; no fleet-scale, multi-building, or environment-general claim. |

**Supporting-evidence note.** The controlled offline comparisons, model-substitution replay and matcher/guard diagnostics support the reasoning and memory contributions (C2/C3); they are not separate contributions. Sessions A–C remain development-stage evidence.

diagram. The confirmatory validation campaign of the frozen v4.8-review system, Session E, is detailed in Section 6. Table 2 therefore separates the inherited basis from the elements introduced in this work and identifies the evidence and claim scope associated with each revised contribution.

Table 2 also establishes the claim ceiling used in the remainder of the manuscript: direct experimental support is strongest for the frozen higher-layer reasoning mechanisms and M3 memory lifecycle, while the broader M1–M5 taxonomy and the inherited L0–L2 foundation have different evidentiary roles. Sections 4 and 5 describe the mechanisms within those boundaries; Sections 6 and 7 define and report the corresponding experimental evidence.

## 4. Hybrid semantic reasoning

Layer 3 bridges operator intent expressed in natural language and an executable graph-grounded navigation target. Its deterministic-first path, L3a, uses normalized language, graph structure, semantic-object context and promoted instruction–node associations. L3b provides VLM-based visual-semantic resolution when no deterministic candidate is accepted or a safeguard requires escalation. The following description specifies this architecture's frozen v4.8-review implementation.

The separation is operational rather than a claim that one path is universally sufficient. L3a is intended for cases that can be resolved under explicit rules and acceptance conditions; L3b is reserved for cases that require richer semantic interpretation or guarded resolution. The two paths share a common action-validation and audit boundary, and post-arrival confirmation is treated as a downstream outcome distinct from initial target selection and physical navigation completion. Fig. 2 summarizes the frozen control flow, while Algorithm 1 specifies the deterministic resolver and its escalation conditions.

### 4.1. Design rationale and executive contract

The deterministic-before-VLM ordering is motivated by three operational requirements that remain valid from the submitted design. First, many indoor navigation requests refer directly to graph nodes, known objects, object classes, explicit attributes, or previously learned instruction–node associations. Such cases do not require open-ended visual-language inference when a unique admissible destination can be established from the structured context. Second, VLM inference is substantially more computationally expensive and temporally variable than lexical and graph-based matching, so invoking it indiscriminately would add avoidable latency and consume the shared inference resource. Third, a rule-based front end provides an explicit accept-or-defer boundary: if its conditions are not met, the system can continue the cascade or escalate rather than forcing a weak partial match.

L3a is a deterministic semantic resolver whose acceptance rules use graph relations, semantic objects, pose where needed and compiled memory. It does not reproduce general VLM reasoning; affordance-level or otherwise unresolved interpretations remain candidates for L3b.

The reasoning layer is bounded by the executive-contract formalization $\langle A, O, V, L \rangle$ adopted from ROSClaw [10] and specialized here for semantic navigation. A denotes the affordance space exposed to the reasoning layer, including graph-grounded destinations and admissible navigation actions. O denotes the normalized observation and context presented to the resolver or model-facing interface. V denotes the validation boundary applied before an accepted semantic proposal can become a robot action. L denotes structured audit logging that preserves the instruction, resolution path, validation state, and downstream outcome. The contract is an interface and safety boundary; it is not itself a semantic reasoning algorithm.

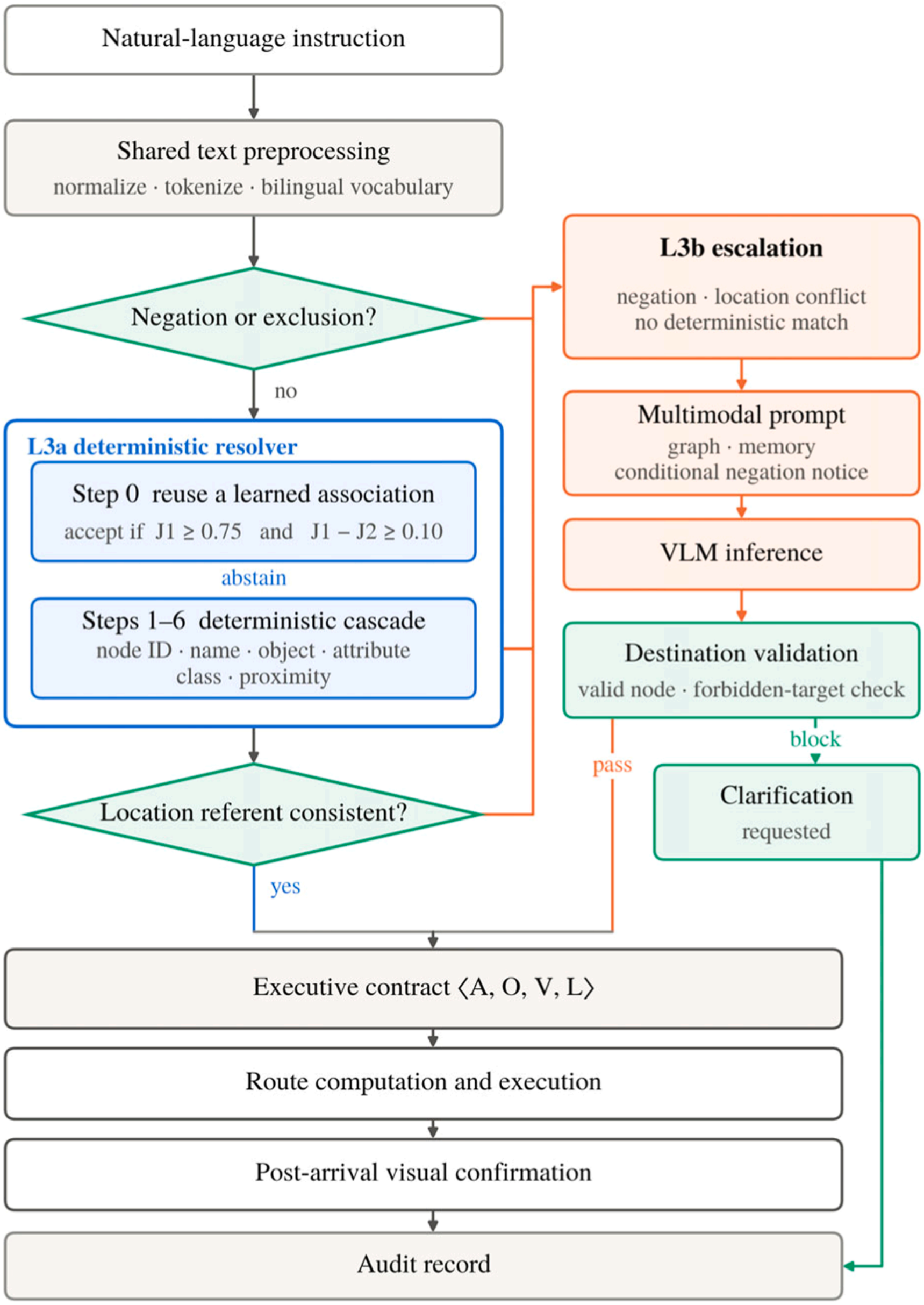

**Fig. 2.** Frozen v4.8-review reasoning flow. Detected negation/exclusion precedes all resolver steps. Otherwise, Step 0 accepts an M3 association only when the similarity threshold and runner-up margin are met; abstention continues to Steps 1–6 rather than directly escalating. Location conflicts trigger escalation, except for explicit numeric-node commands. Negation, location conflict and unresolved deterministic selection converge on L3b, followed by destination validation; a blocked answer is logged without navigation. Algorithm 1 specifies the resolver. A frame is acquired before resolution on every cycle: deterministic decisions avoid VLM inference, not frame acquisition. Model-call and parsing failures are separate outcomes omitted from the diagram.

This architecture deliberately separates semantic selection from physical execution. A successful L3 decision produces an admissible target or an explicit non-executable outcome such as escalation or clarification. Nav2 execution remains downstream at L1, and post-arrival confirmation is evaluated separately in Section 4.7. Consequently, a target-selection decision, a navigation completion, and a confirmation result are distinct events in the audit trail.

The computational boundary must also be distinguished from the surrounding orchestration. In the frozen Session E implementation, the pre-navigation image available for a possible L3b call is acquired before the deterministic resolver timer starts. L3a computation itself neither consumes that image nor invokes the VLM, but the sub-millisecond resolver timing reported later is therefore resolver-compute latency rather than end-to-end instruction latency, and it must not be interpreted as evidence that no image was acquired during that decision cycle.

Separately, a live image-request transaction was measured on Xplorer-B from the workstation running the navigator, using the deployed sequence of an HTTP image request, JSON decoding and extraction of the base64 image string. The interval ended when the image string was available in memory for L3b; JPEG decoding and validation occurred after the timer. Fifty transactions were specified before measurement, with a one-second idle gap, no retry or discarded transaction, and the route graph and navigator commit matching Session E. All 50 succeeded, with a median of 57.29 ms (Q1–Q3 46.43–68.17 ms; range 34.75–211.53 ms). The first transaction,

**Algorithm 1**

Frozen v4.8-review L3a deterministic resolver. A negation/exclusion guard precedes every resolution step; Step 0 reuses a promoted instruction–node association under an acceptance threshold and a runner-up abstention margin; Steps 1–6 apply the deterministic cascade over the frozen semantic map. A location-consistency check escalates any resolution that contradicts an explicit location identifier in the instruction, except at Step 1, where the identifier is itself the resolution. Abstention at Step 0 is not an escalation: control falls through to Step 1.

```
Require:  instruction u; node set N; object-node map O; robot pose p;
          installed association index D
Require:  acceptance threshold θ = 0.75; abstention margin δ = 0.10
Ensure:  (n, s) - destination node n resolved at cascade step s -
          or ESCALATE(r), handing the decision to L3b with reason r

⟨negated, M⟩ ← DetectNegation(u)
▷ guard: precedes every step
if negated then return Escalate(negation_detected, M)
X ← ExplicitLocationNodes(u, N, O)
▷ nodes named by an explicit identifier; ∅ if none
ū ← Normalize(ExpandAliases(u))

▷ Step 0 - reuse of a promoted instruction–node association
if D ≠ ∅ then
    T ← Tokenize(u)
    if T ≠ ∅ then
        for each destination node m referenced by D do
            σ[m] ← max over stored examples e of m of Jaccard(T, Tokenize(e))
            ▷ aggregated per node, not per example
        ⟨n1, J1⟩, ⟨n2, J2⟩ ← the two highest-scoring distinct nodes of σ
        if J1 ≥ θ and n1 ∈ N and (n2 undefined or J1 − J2 ≥ δ) then
            if X ≠ ∅ and n1 ∉ X then return Escalate(location_conflict)
            return (n1, 0)
            ▷ else abstain and fall through to Step 1

▷ Steps 1–6 - deterministic cascade over the frozen semantic map
⟨n, s⟩ ← ⟨⊥, ⊥⟩
if ū names an explicit node index k and k ∈ N then
    ⟨n, s⟩ ← ⟨k, 1⟩
else if exactly one node name in N matches ū then
    ⟨n, s⟩ ← ⟨that node, 2⟩
else if exactly one object identifier in O matches ū then
    ⟨n, s⟩ ← ⟨its nearest node, 3⟩
else if attribute scoring over O has a unique maximum then
    ⟨n, s⟩ ← ⟨its nearest node, 4⟩
else if exactly one object class C is named in ū then
    Q ← objects of class C
    if Q contains surveyed objects then Q ← surveyed objects only
    if |Q| = 1 then
        ⟨n, s⟩ ← ⟨nearest node of the single object, 5⟩
    else if ū carries a proximity cue and p is available then
        ⟨n, s⟩ ← ⟨nearest node of argmin_{q∈Q} dist(p, q), 6⟩

if n = ⊥ then return Escalate(no_deterministic_match)
if s ≠ 1 and X ≠ ∅ and n ∉ X then return Escalate(location_conflict)
▷ Step 1 is exempt: the explicit identifier is the resolution

return (n, s)
```

211.53 ms, was retained. This describes the request-to-image-availability transaction, including its HTTP transfer and client decoding, rather than separately timing sensor exposure or establishing frame age. It is reported separately and is not added to the offline decision latencies in Sections 6.8 and 7.7.

### *4.2. Input normalization and semantic context*

L3a and the guard logic operate on a shared bilingual text-processing layer so that the same linguistic conventions are applied to direct graph matching, M3 association matching, and safeguard detection. The frozen sas_text.py v1.2.0 processor lowercases and trims the instruction, applies Unicode NFKD decomposition to remove diacritics, replaces underscores with spaces, collapses repeated whitespace, normalizes apostrophes, and removes punctuation for word-level processing. Romanian class aliases are expanded before M3 content-token filtering so that semantically equivalent class references can enter the same deterministic vocabulary.

Two related text representations are retained because they serve different purposes. The raw normalized wording is used for negation and exclusion detection before stopword removal, preventing a negative marker from disappearing during content-token filtering. The M3 representation is a set of content tokens with a minimum token length of two characters after stopword removal and alias expansion. The frozen text resource contains 116 stopwords, 41 token-level negation markers, 15 negation phrases, 9 strong exclusion terms, and 22 Romanian-to-POI-class aliases; the exact lists and processing functions are supplied by the released sas_text.py rather than reproduced as a long vocabulary table in the manuscript.

The semantic context combines the navigation graph, semantic objects, robot state, and installed memory. Graph nodes provide identifiers, names, and coordinates. Semantic objects provide object identifiers, classes, attributes, source information, and their mapping to graph nodes. The deterministic resolver prioritizes surveyed/static objects over dynamic detections for class-based resolution when both are available, thereby avoiding a dynamic duplicate from displacing an already established static referent. Current pose is used only by rules that require spatial proximity.

Object and attribute interpretation remains deliberately lexical on the deterministic path. Explicit object identifiers and node names are matched against their normalized forms. Attribute matching uses word-bounded lexical matches and resolves only when one object attains a unique best attribute match. The resolver does not infer an unstated affordance merely because an object possesses a semantically related attribute. Requests requiring that broader interpretation can therefore remain unresolved by L3a and proceed to L3b.

Location information is treated separately from generic semantic attributes. The safeguard layer extracts explicit alphanumeric location referents, such as room or laboratory identifiers, and maps them to graph nodes through node names and object identifiers. These referents constrain candidates produced by Step 0 and Steps 2–6, as detailed in Section 4.5. An explicit numeric graph-node command in Step 1 is handled as a direct node reference after the negation guard and is therefore exempt from the alphanumeric location-conflict test.

### 4.3. Seven-step deterministic resolver

The frozen v4.8-review L3a resolver is a first-resolving-step cascade. It receives the instruction, graph nodes, semantic-object-to-node context, current pose when available, and the installed M3 association index. Before any deterministic step is evaluated, the negation/exclusion guard is applied. If that guard does not require escalation, the resolver evaluates Steps 0–6 in order and returns as soon as one step produces an admissible candidate that survives the applicable conflict checks. If no step yields an admissible deterministic resolution, the decision is handed to L3b.

The seven steps preserve the structure of the submitted resolver while incorporating the final safeguards and acceptance rules. Step 0 attempts reuse of a promoted M3 instruction–node association under the threshold and runner-up margin described in Section 4.4. Step 1 recognizes an explicit numeric graph-node identifier. Step 2 requires a unique normalized graph-node-name match. Step 3 requires a unique semantic-object-identifier match. Step 4 requires a unique best lexical match over eligible object attributes. Step 5 resolves a single mentioned object class only when the filtered candidate set contains one object. Step 6 handles a single class with multiple candidate objects when a proximity cue is present and a current pose is available, selecting the nearest candidate.

For Steps 5 and 6, surveyed/static objects take precedence over dynamic detections when at least one surveyed object of the matched class is available. For Steps 0 and 2–6, an accepted candidate is still subject to the explicit-location consistency check when the instruction contains an alphanumeric location referent. The direct numeric node command of Step 1 is the only deterministic candidate exempt from that later location check, although it remains downstream of negation/exclusion precedence.

Fig. 2 and Algorithm 1 serve complementary purposes. Fig. 2 shows the complete L3a/L3b control flow, including the guard order and downstream execution boundary. Algorithm 1 provides the procedural detail needed to reproduce the deterministic cascade, including the Step-0 ranking rule, first-resolving-step semantics, location-conflict checks, and escalation outcomes. The two assets use the same frozen ordering and parameters.

The resolver is deterministic conditional on the same instruction, graph and semantic-object context, pose where required, and installed memory state. This property should not be read as a universal fast-path guarantee: which step resolves a request depends on the workload and available context. The observed first-resolving-step composition and timing are therefore reported as Session E results rather than as intrinsic performance constants of the algorithm.

### 4.4. Step-0 M3 matching, ranking, threshold, and abstention

Step 0 provides the deterministic reuse interface between the M3 memory lifecycle of Section 5 and the L3a cascade. Each promoted M3 entry contains one or more normalized instruction examples associated with a destination node. For an incoming instruction, the shared tokenizer produces a content-token set T. Each stored example e is tokenized in the same way, and lexical overlap is measured by the Jaccard coefficient

$$J(T, e) = |T \cap e| / |T \cup e|$$

Because multiple stored examples may point to the same destination, Step 0 ranks destination nodes rather than individual examples. For each node, only the maximum example-level Jaccard score is retained. The best-scoring node therefore has score J1, while J2 denotes the score of the best competing destination node, if a competitor exists. Examples that support the same node reinforce its candidate set without acting as runner-up competitors to one another.

The frozen v4.8-review acceptance rule uses a Jaccard threshold of 0.75 and a runner-up margin of 0.10. The top node is accepted only when J1 is at least 0.75 and either no runner-up destination exists or $J1 - J2$ is at least 0.10. This rule separates two failure modes: insufficient lexical support and a near-tie between competing destinations. A tie or near-tie is therefore not converted into an arbitrary best guess.

Step-0 rejection is local to the memory-matching step. If the threshold or margin condition is not met, the resolver continues to Steps 1–6; it does not immediately invoke the VLM. Conversely, a Step-0 candidate that satisfies the numerical acceptance rule can still be rejected by the explicit-location consistency safeguard. If the instruction names an alphanumeric location referent that is incompatible with the candidate node, the resolver escalates the decision as a location conflict rather than executing the stored association.

The threshold and margin are fixed properties of the frozen implementation described and validated in this manuscript. They are not presented as globally optimal hyperparameters. The post-hoc Step-0 robustness characterization reported later examines the observed decisions under alternative threshold and margin values; it was not used to choose or tune the frozen settings.

### 4.5. Negation precedence and location-conflict handling

The final v4.8-review resolver places explicit safeguards around deterministic reuse rather than treating them as incidental parser exceptions. The first safeguard is negation/exclusion precedence.

Detection is performed on the raw normalized instruction before stop-word filtering and before any deterministic Step 0–6. When a negation or exclusion pattern is detected, the deterministic cascade is skipped and the instruction is handed to a negation-aware L3b path. Thus, a previously promoted association or even an otherwise valid explicit node reference cannot override an exclusion expressed in the instruction.

The negation-aware handoff also constructs a forbidden-target constraint for the model-facing path. The orchestration removes the negative markers from the relevant wording to identify candidate targets that the operator is excluding, supplements that set with applicable class/static-object evidence, and preserves explicit location referents. The L3b prompt then carries an explicit negation notice together with the forbidden-node set. A model response that proposes no usable node, returns the clarification value $-1$, cannot be verified against the forbidden set, or selects a forbidden destination is non-executable and is blocked from navigation; only a non-forbidden, graph-valid destination may continue to the downstream action validator.

The second safeguard addresses contradictory location evidence. Explicit alphanumeric referents such as room identifiers are mapped to a set of compatible graph nodes before semantic matching. If Step 0 or Steps 2–6 produce a candidate outside that set, the resolver escalates the decision as a location conflict rather than allowing a semantically plausible but location-inconsistent target to pass. This check is intentionally separate from the direct numeric-node rule: Step 1 is treated as an explicit graph-node command and is exempt from the alphanumeric location-consistency guard, but it remains subordinate to the initial negation check.

These safeguards define precedence and escalation behavior, not a general safety certification. They reduce two concrete failure modes relevant to memory reuse and semantic target selection, while leaving unresolved or conflicting language to the deliberative path. Their controlled behavior under Session E is reported in Section 7 rather than embedded in the method definition.

### 4.6. L3b application interface and action validation

L3b is entered when no admissible deterministic resolution is available or when a safeguard requires escalation, such as for negation/exclusion or a location conflict. Its role is not to emit arbitrary robot control commands. Instead, it performs constrained semantic destination selection through a fixed multimodal application interface, after which the proposed destination is parsed and validated before any downstream navigation action is allowed.

The model input combines the current instruction with the graph-grounded context already available to SAS and the pre-navigation image associated with the decision cycle. The deployed prompt is organized around six functional blocks inherited from the submitted implementation and retained in the frozen interface: the installed memory/context prefix when present; the current robot state; the navigation graph with node identifiers, names and coordinates; semantic objects grouped by graph region with relevant attributes and source information; designated seating/resting information where encoded by the ontology; and the task plus explicit response and navigation rules. Negation-aware escalations add the notice and forbidden-target information described in Section 4.5.

The L3b response contract requests a machine-readable object containing an integer destination-node identifier and a textual reason. The deployed parser strips Markdown fences and parses after a closing think tag when a reasoning preamble is present despite the disabled-thinking request. Parsed proposals are checked against the graph and active safeguards; explicit $-1$, invalid nodes and safeguard violations do not produce navigation commands. The parser does not strictly check node-identifier types, however; Section 7.8 reports the boolean-input weakness. Missing-answer and parsing failures remain distinct from deliberate semantic clarification.

The executive contract introduced in Section 4.1 supplies the common boundary around this model-facing operation. The affordance component exposes graph-grounded targets and admissible navigation actions; the observation component supplies the normalized state and semantic context; the validator checks the parsed proposal against the graph and configured safety constraints; and the logger records the reasoning path, proposed and accepted target state, timing fields, and downstream execution/confirmation outcome. This organization keeps target interpretation separate from the Nav2 action interface and makes non-executable model outputs explicit in the audit trail.

The frozen Session E control configuration used the pretrained qwen3.5:4b model through Ollama. Its model digest, quantization, runtime, software, and frozen configuration identity are consolidated in Section 6.3 and Table 3 rather than repeated here.

**Table 3**
Platforms, compute topology and frozen v4.8-review identity. Part A describes Xplorer-B/C and robot–workstation partitioning; Part B identifies the Session E software, model, parameters and provenance. Shared sensing/map resources and differing onboard hardware are listed explicitly.

Part A - deployment and compute topology

| | Xplorer-B | Xplorer-C |
|---|---|---|
| Architecture label | single-rpi5 | dual-rpi5 |
| Navigation compute | Raspberry Pi 5, 16 GB, ARM Cortex-A76, Ubuntu 24.04 | same |
| Detector compute (YOLO) | same board as navigation | second Raspberry Pi 5, dedicated, over Ethernet |
| Camera | Logitech C920 HD, 640 × 480, USB | same |
| 2D LiDAR | STL-19P (LD19), 360°, ≈503 rays, 10 Hz | same |
| IMU | BNO055, 9-DoF, I²C | not fitted |
| Drivetrain | 4WD differential, quadrature encoders | 4WD differential, hall encoders |
| Motor control | 2 × RoboClaw 2 × 15 A, UART | 2 × L298N + ESP32 |
| Battery telemetry | front/rear battery-state topics | not instrumented equivalently |
| Context server / route graph | per-robot context server; canonical route graph | per-robot context server; same canonical route graph |

**Layer placement is identical on both robots: L0–L2 run onboard; L3a, L3b and L5 audit logging run on the workstation over HTTP. A GPU is used only for L3b inference. Onboard Nav2 receives goals through ROS 2 action interfaces.**
**The shared workstation (Intel Core Ultra 7 255H, NVIDIA RTX 5050, 32 GB RAM, Ubuntu 24.04) hosts SAS/VLM services, audit logging and RViz. ROS 2 Jazzy was used throughout. Workstation hardware was common to Sessions A–C/E; software identity follows the campaign boundary in Table 4.**

Part B - frozen configuration and provenance

| Field | Value |
|---|---|
| Navigator version | v4.8-review |
| Commit | f973aa0c4920e47cdd85d7e4e263e3b66fb41ddc |
| Working tree | clean at every one of the 214 recorded startups |
| Protocol version | E-2026–09 |
| Audit schema provenance tag | audit_schema_version = 2.1* |
| VLM | qwen3.5:4b, digest 0c8faadc…5018, quantization Q4_K_M |
| Model adaptation | pretrained VLM; no task-specific model fine-tuning or separately trained task classifier |
| Runtime | Ollama 0.30.2; NVIDIA driver 580.173.02 |
| Text-processing module | sas_text 1.2.0 |
| Step-0 acceptance threshold | **0.75** |
| Step-0 runner-up abstention margin | **0.10** |
| Promotion rule | frequency ≥ 3 **and** rounded consistency ≥ 0.80 **and** ≥ 1 accepted L3b VLM resolution |
| Semantic-object ontology | md5 2e483750…97c5 |
| Route graph | md5 e0130b78…9817, identical from both context servers |

* All 214 startup records also contain audit_schema = 2.2; the frozen configuration record fixes audit_schema_version = 2.1 as the provenance identity. Both fields are retained here without inferring an undocumented equivalence.

The fixed prompt/parser contract is intended to make model substitution possible at the application-interface level, but the architecture does not assume that different VLMs are semantically equivalent. The controlled replay reported later evaluates whether alternative models can operate through this same interface without model-specific prompt or parser changes. Any such result is interpreted as portability at the L3b application interface, not semantic model invariance.

### 4.7. *Post-arrival confirmation*

Target selection and physical arrival do not by themselves establish that the intended semantic referent is observable at the destination. The SAS therefore treats post-arrival confirmation as a separate downstream decision after the navigation action has completed. This preserves a distinction that was present in the submitted implementation but is made explicit in the revision: semantic target resolution determines where the robot should attempt to navigate, Nav2 reports the physical execution outcome, and confirmation evaluates the available evidence at arrival.

The confirmation route is selected from the destination's configured confirmation metadata. For landmark and contextual signatures, the system evaluates an arrival image against the expected semantic context and records a structured confirmation result, including the confirmation Boolean and confidence level. For destinations handled without a visual confirmation signature, the system uses its configured pose-based confirmation path rather than the VLM-based confirmation path.

The confirmation record is stored independently from the navigation outcome. A completed navigation action can therefore be followed by a negative or low-confidence visual confirmation without being silently rewritten as a navigation failure, and a pose-based confirmation does not imply that a visual referent was actually observed. This separation also prevents aggregate confirmation prevalence from being interpreted automatically as semantic accuracy.

The controlled present/absent evaluation and the broader descriptive confirmation observations are reported in Section 7.3. Those results characterize the implemented confirmation logic under the tested conditions; they do not turn post-arrival confirmation into a general scene-recognition guarantee.

## 5. Scoped semantic memory

Layer 5 converts selected audit history into structured M1–M5 memory and a compact runtime digest. In the frozen implementation, M3 stores learned instruction–node associations inferred from repeated model-mediated decisions; these are not user-confirmed preferences. Shared-memory reuse across robot platforms proceeds through deliberate digest compilation and installation. Section 5.7 distinguishes the implemented memory organization from its experimentally isolated effects.

Fig. 3 summarizes both parts of the mechanism. Panel (a) defines the

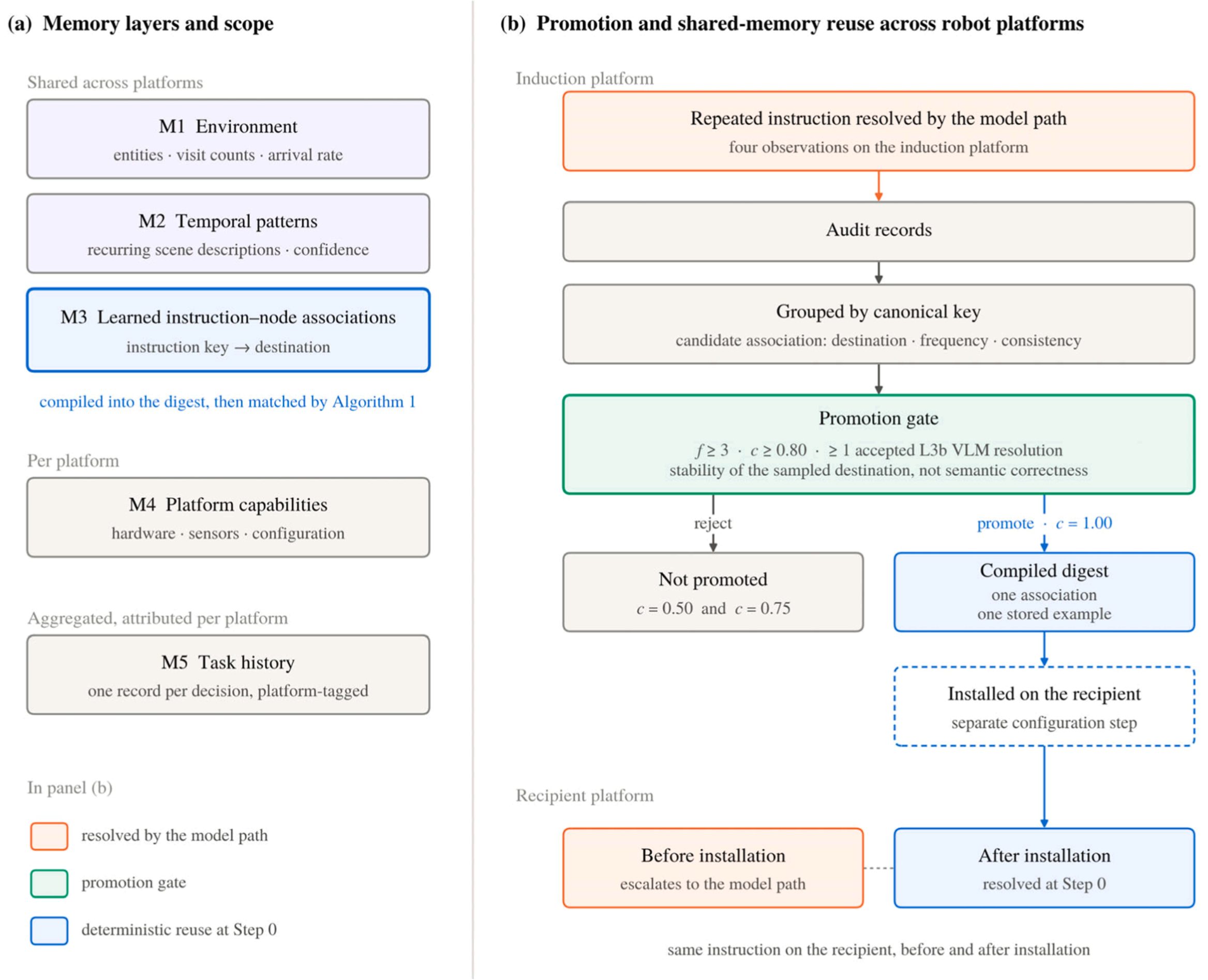

**Fig. 3.** Scoped memory and shared-memory reuse. (a) M1–M3 are shared across platforms; M4 is platform-specific; M5 aggregates task history with per-platform attribution. Only M3 is matched at Step 0. (b) Audit-derived associations are grouped by canonical key and promoted at frequency $\geq$ 3, dominant-destination consistency $\geq$ 0.80 and at least one accepted L3b resolution. This gate measures sampled-output stability, not semantic correctness. The illustrated candidates have consistencies 0.50 and 0.75 (rejected) and 1.00 (promoted). The promoted digest contains one association and one stored instruction. Its deliberate installation changes recipient-platform resolution from L3b to Step 0. Algorithm 2 specifies extraction and promotion.

**Algorithm 2**

Frozen v4.8-review audit-to-M3 extraction and promotion. Decision records are filtered from a selected audit subset, grouped under a canonical cluster key built from the same frozen preprocessing Algorithm 1 tokenizes with, and passed through a three-condition promotion gate. Promoted associations are compiled into a digest that Algorithm 1 can consume as its Step-0 index. The gate tests the stability of a sampled output, not its semantic correctness.

```
Require:  audit records A of a selected campaign subset
Require:  auxiliary compiled-memory summaries S - environment, temporal patterns,
          platform capabilities, task history
Require:  promotion frequency ϕ = 3; consistency floor γ = 0.80; compaction trigger B
Ensure:  candidate associations P and compiled digest D; D is installable as the
          Step-0 association index of Algorithm 1

▷ Stage 1 - filter
    A_d ← the records of A typed as decisions
    ▷ startup and other events are set aside
    attribute every record to its platform through its session's startup event

▷ Stage 2 - group
    for all d ∈ A_d do
        k ← ClusterKey(instruction(d))
        ▷ canonical key from the same frozen preprocessing Algorithm 1 tokenizes with
        append ⟨instruction(d), destination(d), method(d), time(d)⟩ to group[k]

▷ Stage 3 - promotion gate
    P ← ∅
    for all k such that group[k] ≠ ∅ do
        f  ← |group[k]|
        ▷ frequency
        n* ← the most frequent destination in group[k]
        c  ← round( count(n*, group[k]) / f , 2)
        ▷ consistency, two decimals
        v  ← the number of records in group[k] whose destination
           the VLM path resolved and the system accepted
        E  ← the first three distinct instructions of group[k], in lexicographic order
        emit association ⟨k, n*, f, c, E⟩, promoted ⇔ ( f ≥ ϕ ∧ c ≥ γ ∧ v ≥ 1 )
        if promoted then P ← P ∪ { ⟨n*, f, E⟩ }

▷ Stage 4 - compile
    D ← ⟨ S, { ⟨n*, f, first(E)⟩ : ⟨n*, f, E⟩ ∈ P } ⟩
    if size(D) > B then
        compact the environment and temporal summaries in S once
        ▷ the promotion set is never reduced, and the size is not re-tested
    stamp D with a content digest over its stable fields, so that two compilations
        from the same records carry the same identity
    write the candidate associations and D
    return P, D
```

M1–M5 memory layers and their deployment scope; panel (b) shows the controlled audit → association → promotion/rejection → digest compilation → installation → Step-0 reuse workflow. Algorithm 2 specifies the frozen audit-to-M3 extraction and promotion procedure. Section 5 therefore distinguishes the memory taxonomy from the specific M3 promotion mechanism and from the experimental evidence reported later.

### 5.1. M1–M5 taxonomy and scope

The M1–M5 organization separates information that can be shared across robot platforms from information that must remain platform-aware. This scope distinction is central to the design: information about the environment and instructions can be useful beyond the session or platform that produced it, whereas capability and task-history records must retain platform attribution. Fig. 3(a) shows the resulting taxonomy. Fig. 3 emphasizes the platform-sharing axis because it is the one directly exercised by the present experiments; map scope and operator scope remain orthogonal partitioning dimensions of the framework where applicable and are not evaluated independently here.

M1 - Environment. M1 summarizes operational information associated with the environment and its entities, including visit-related statistics and arrival-related information. Its purpose is to preserve environment-level context that can be reused across platforms operating against the same semantic environment.

M2 - Temporal patterns. M2 stores recurring scene-level descriptions and associated confidence information derived from operational observations. Like M1, it contributes contextual information that can be shared across platforms when the underlying environment is common.

M3 - Learned instruction–node associations. M3 represents recurring mappings between a normalized instruction key and a destination node. It is the memory layer connected directly to deterministic Step 0: promoted associations are compiled into the digest and can subsequently be matched by Algorithm 1 before L3b inference. Where the framework associates such mappings with an operator scope, that scope identifies how memory may be partitioned; it does not establish user-ground-truth preference validity for an experimentally inferred mapping.

M4 - Platform capabilities. M4 records platform-specific capability and configuration information, including hardware, sensing, and configuration state. It is retained per platform rather than merged into a single cross-platform capability record.

M5 - Task history. M5 records decision-level operational history with platform attribution. In the frozen implementation it is aggregated while preserving which platform produced each record, allowing cross-session analysis without erasing platform identity.

The compact scope labels in Fig. 3 should therefore be read as deployment rules rather than as claims of equal experimental importance. M1–M3 are represented as shared across platforms; M4 is per platform; M5 is aggregated with per-platform attribution. The runtime roles are also asymmetric: M1/M2 can contribute contextual prompt content, whereas M3 can alter the resolution path directly by enabling deterministic Step-0 reuse. The empirical scope of these distinctions is stated explicitly in Section 5.7.

### 5.2. Audit-to-association extraction

Memory extraction is an offline operation over an explicitly selected audit subset rather than a hidden success-only filter. The selected records are first separated by event type so that decision records form the observation population used for association extraction, while startup and other events are retained for provenance and platform attribution where needed. Platform identity is recovered from the corresponding session startup record, but it is not part of the M3 instruction-grouping key; observations from different platforms can therefore contribute to the same candidate association when they belong to the selected extraction scope.

Each retained decision contributes its instruction, destination outcome, resolution path, and time information to an instruction group. Repeated decision cycles are deliberately retained because they define the observed frequency and destination consistency of the candidate. They are not deduplicated into one trial. Only the representative instruction strings stored with an association are deduplicated.

The grouping key is constructed from the same frozen text-processing basis used by Algorithm 1 at match time. Extraction and Step-0 matching are not the same operation: extraction reduces an instruction to a canonical set-valued key for grouping, whereas Algorithm 1 tokenizes stored and incoming examples and scores their similarity. They nevertheless share the normalization, tokenizer, stopword list, and bilingual alias conventions described in Section 4.2. Negation markers are not removed as stopwords in this shared vocabulary, so a negated instruction does not collapse into the same grouping key as its affirmative counterpart.

This procedure produces candidate instruction–node associations; it does not by itself decide that a candidate is reusable. Promotion is a separate gate applied after grouping, as described in Section 5.3 and Algorithm 2.

### 5.3. M3 grouping and promotion rule

For each non-empty instruction group, the extractor computes three quantities used by the frozen promotion rule. Frequency f is the number of retained decision observations in the group. The dominant destination is the most frequently observed destination outcome. Consistency c is the dominant-destination count divided by f and rounded to two decimal places. Null destination outcomes remain in the destination accounting and in the denominator; they are not silently removed from an unstable group.

The third quantity records whether the group contains at least one destination that was resolved through L3b and accepted by the system. Clarification outcomes with no accepted destination and error or parse escalations do not satisfy this condition. The distinction is important because the promotion rule is intended to turn repeated model-mediated semantic outcomes into a reusable deterministic association, not to duplicate instructions that the deterministic cascade already resolves without such an observation.

A candidate is promoted only when all three frozen conditions hold: $f \geq 3$, $c \geq 0.80$, and at least one accepted L3b VLM resolution.

Representative instruction strings are retained separately from the repeated observation count: the frozen procedure keeps up to three distinct instructions per candidate in lexicographic order. Algorithm 2 gives the complete filtering, grouping, frequency/consistency calculation, accepted-L3b gate, promotion decision, and digest-compilation sequence, and it preserves the two-decimal rounding convention used by the frozen implementation.

The extraction procedure and Algorithm 1 share preprocessing but have different roles. Algorithm 2 creates and promotes candidate associations; Algorithm 1 consumes the installed promoted index at Step 0. Step-0 Jaccard ranking, the 0.75 acceptance threshold, the 0.10 runner-up margin, negation precedence, and location-conflict handling therefore remain properties of Section 4 rather than being duplicated in the promotion algorithm.

### 5.4. Promotion stability versus semantic correctness

The promotion gate is a sampled-output stability criterion, not a semantic-correctness criterion.

Its three conditions ask whether a recurring instruction has been observed often enough, whether one destination dominates the sampled outcomes strongly enough, and whether at least one accepted outcome came through the model path. None of those conditions checks whether the dominant destination is semantically correct, whether it matches a user's intended preference, or whether another admissible destination would be better. Accordingly, a single predefined destination should not be interpreted as universal semantic ground truth for inherently ambiguous language; where the controlled evaluation permits multiple plausible destinations, the revised scoring uses predefined acceptable-node sets rather than a single forced target. The submission-era “sit and relax” case illustrates this boundary: a chair and a radiator-area destination can both be semantically plausible, so that observation is retained only as an example of the limitation of single-target scoring and not as evidence of a user-confirmed preference.

This distinction is consequential because promotion moves an association into a path that can later bypass L3b at Step 0. A stable but semantically wrong association can therefore satisfy the gate just as a stable and correct association can. The frozen negation and location-conflict safeguards described in Section 4 constrain important downstream failure modes, but they do not convert the promotion rule into a semantic-validity test.

The later robustness analyses are interpreted under this same boundary. They characterize how low-sample promotion decisions depend on the observed outputs and, in the controlled replay, on model/sampling behavior. They are not treated as evidence that the gate discovers semantic truth or user-confirmed preference validity.

### 5.5. Digest compilation, installation, and Step-0 reuse

Promoted M3 associations are compiled together with the auxiliary operational-memory summaries into a runtime digest. In the frozen procedure, each promoted association contributes its destination, observed frequency, and one stored instruction example to the compiled Step-0 association index. The resulting digest also carries the auxiliary memory summaries used by the higher-layer context path. The compilation stage therefore connects offline audit processing to both deterministic reuse and memory-conditioned model context without making the two mechanisms equivalent.

The digest uses a compact representation, but the size policy is deliberately described as a soft control rather than as a hard token guarantee. The frozen implementation applies a one-pass compaction trigger at approximately 2000 serialized characters, corresponding only approximately to 500 tokens. If that trigger is exceeded, the environment and temporal summaries are compacted once. The promoted association set is not reduced by this branch and the serialized size is not tested again. Consequently, the mechanism limits prompt growth in the common case but does not guarantee an exact 500-token upper bound.

Compilation and installation are separate operations. Algorithm 2 creates and writes the candidate-association output and compiled digest; it does not automatically install that digest on a robot platform. Installation is a deliberate configuration step that selects which compiled memory state a subsequent navigator run will load. This separation makes the memory intervention auditable and prevents a new association from becoming active merely because the extraction process was executed.

Once an installed digest is loaded, its promoted M3 index becomes available to Algorithm 1 at Step 0. The incoming instruction is then matched under the threshold, margin, and safeguard rules defined in Sections 4.4 and 4.5. If an association is accepted, the instruction can be resolved deterministically; if it is rejected or conflicts with the current instruction, the remaining deterministic cascade or L3b path remains available.

### 5.6. Shared-memory reuse across robot platforms

The cross-platform mechanism is shared-memory reuse across robot platforms through a centrally managed compiled artifact. One platform can contribute observations to the selected audit history during an induction phase; after offline grouping and promotion, a newly compiled digest can be installed for a recipient platform. If the same instruction is later issued on the recipient and the installed M3 entry satisfies the Step-0 matching and safeguard conditions, Algorithm 1 can reuse that association deterministically.

Fig. 3(b) makes the phase boundary explicit. The induction platform produces repeated model-mediated observations; the audit records are grouped and passed through the promotion gate; a qualifying association is compiled into the digest; and installation on the recipient occurs as a separate configuration step. The before/after recipient comparison therefore changes the available compiled memory state deliberately rather than through autonomous inter-robot exchange.

Accordingly, the mechanism is not direct robot-to-robot learning. The recipient does not receive an online update directly from another robot, and no autonomous peer-to-peer learning protocol is claimed. The transferable object is the compiled memory artifact, whose shared layers can be loaded by another platform while platform-aware records retain their attribution. The physical evidence for this mechanism is intentionally reported later at the level of one induced association and repeated recipient reuse observations rather than as an independent-association transfer rate.

### 5.7. Experimental scope of M1–M5

The M1–M5 taxonomy is broader than the set of memory effects that can be isolated experimentally in this revision, and that asymmetry is part of the claim boundary.

M3 is the directly demonstrated adaptive mechanism. The confirmatory memory experiments exercise candidate promotion/rejection, digest compilation and installation, deterministic Step-0 reuse, and shared-memory reuse across robot platforms. The promotion-robustness analyses likewise concern M3 rather than the complete taxonomy.

M1 and M2 are not independently evaluated. The compiled-memory removal experiment jointly removes the M3 Step-0 index and the M1/M2 context supplied to L3b, so changed destinations cannot be attributed uniquely to one memory layer. The controlled A–B comparison in Sections 6.8 and 7.7 instead disables M3 target reuse while retaining M1/M2 context and common safeguards, supplying conditional M3 attribution for the five familiar items.

M4 and M5 are implemented as platform-capability and task-history structures and are used for configuration, attribution, provenance, and operational analysis, but they are not independently evaluated for performance contribution. The results therefore do not imply that all five memory layers were separately validated or that each one improves navigation or semantic reasoning.

## 6. Experimental methodology

### 6.1. Platforms, environment, and compute topology

The physical experimental campaigns reported in this paper used two indoor mobile robot configurations, Xplorer-B and Xplorer-C, operating on the same indoor test floor: a corridor and adjoining hall represented by a route graph of 24 nodes and 30 bidirectional connections (60 directed edge records). The test site was the second floor of the Faculty of Industrial Engineering and Robotics (FIIR), National University of Science and Technology POLITEHNICA Bucharest, the same physical environment used in our prior L0–L2 study [5]. The static ontology used in the frozen campaign contains 18 manually annotated semantic objects associated with 15 graph nodes; the remaining nine nodes are topological only. The complete object list, with each object's class, attributes, and nearest graph node, is provided in the public repository as config/semantic_objects_static.geojson. Fig. 4 shows the two platforms, the operational view during a mission, and the route graph with its annotated objects. Both configurations use the same camera/2D-LiDAR sensing stack. During Session E, their context servers exposed the same canonical route graph and frozen semantic-object ontology, whose identities are reported in Table 3.

The two configurations differ in what sits below the semantic layers rather than in the layers themselves. Xplorer-B runs navigation and object detection on one onboard computer and carries an inertial measurement unit; Xplorer-C partitions the two onto separate onboard computers connected over Ethernet and has no inertial unit fitted. They differ further in drivetrain encoders, in motor-control hardware, and in battery-telemetry instrumentation. Table 3, Part A lists these differences field by field. Because power and onboard-compute conditions are not experimentally isolated between the two, any difference observed between the platforms must be read with that caveat; it is not attributable to the semantic reasoning stack or to the navigation algorithm on the evidence reported here. For reference, the inherited L2 implementation reported YOLO26n/NCNN throughput of $5.5 \pm 0.7$ FPS ($167 \pm 21$ ms per inference) on Raspberry Pi 5 [5]; detector latency was not re-estimated as a Session E higher-layer outcome.

The division of work between robot and workstation is identical for both configurations, and it matters for how the reported latencies should be read. Locomotion, sensing and metric navigation run onboard: the metric navigation stack is launched on each robot and commanded remotely over standard action interfaces. Deterministic semantic resolution, model inference and audit logging run on a shared workstation and are reached over HTTP. A graphics processor is required only for model inference; the deterministic resolution path uses none. The workstation also hosts the audit log and the monitoring view. Its hardware was common to the development and confirmatory campaigns, while the SAS software and frozen configuration identity differed according to the campaign boundary reported in Table 4. Each robot has its own context server, while the model inference service is shared, which is the condition one Session E block deliberately exercises by having both robots issue requests to it at the same time.

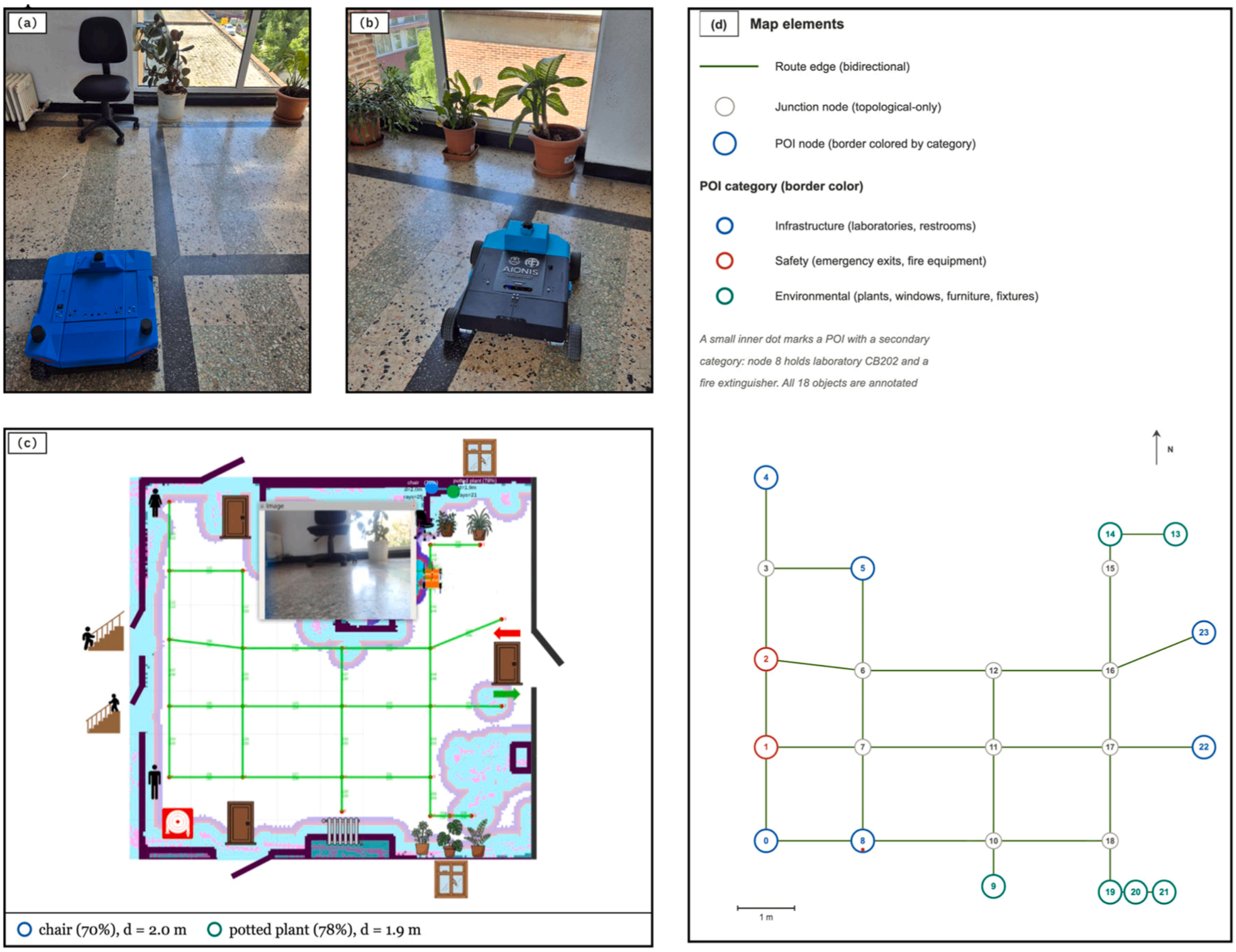

**Fig. 4.** Platforms and operational environment. (a) Xplorer-B and (b) Xplorer-C in the test corridor. (c) Mission view with occupancy map, graph, pose, camera feed and two detections; labels were redrawn for legibility, while the underlying uncropped capture and native overlay were retained. (d) Frozen route graph: 24 nodes, 30 bidirectional edges and 18 annotated objects; nine nodes are topological only. Border colors identify object categories. The panel was generated directly from the graph and semantic-object files served during Session E.

Five wall-mounted fiducial markers, installed for an unrelated experiment, serve only the external localization reference described in Section 6.6; no part of the navigation stack reads them.

### 6.2. Evidence model and versioned experimental campaigns

The evidence comprises development-stage observations from Sessions A–C and the frozen Session E physical campaign (Table 4, Part A). Sessions A–C lack commit-level and clean-working-tree provenance, and Session A also spans two development builds. They are retained in Section 7.9 as development history, not estimates of final-system performance; no Session E parameter or outcome is projected back onto them. Session E is the only physical campaign treated as confirmatory for v4.8-review. Offline analyses provide complementary supporting evidence, while the inherited L0–L2 foundation is documented in Section 3.2. There is no Session D.

Sessions A, B and C contain 10, 3 and 4 audit sessions and 85, 76 and 4 decision cycles, respectively; 37 Session A decisions were scenario-classified. Their Step-0 threshold was 0.60. Table 4 and Section 7.9 report Session A's version-specific counts and the corrections to earlier development estimates. Session E contains 214 audit sessions and 226 decision cycles, with a commit identifier and clean-tree flag recorded at every startup. Its Step-0 threshold was 0.75 with a 0.10 runner-up margin. Section 6.3 and Table 3 consolidate its software, model, graph, ontology and parameter identity.

The confirmatory boundary rests on documented software identity. Session E covers two robot configurations in one indoor environment; its blocks address different questions with different experimental units, so corpus totals are not a pooled experimental N. Section 6.4 and Table 4, Part B define the block design; Section 6.5 defines outcomes, populations and analysis units.

### 6.3. Frozen v4.8-review configuration and provenance

The confirmatory campaign was executed against a single frozen build of the navigator, v4.8-review, and the reproducibility identity of that build is the reference configuration against which every Session E result should be read. Table 3, Part B reports it in full. The identity is not only a version label: it comprises the commit from which the running system was built, the state of the working tree at each start-up, the protocol and audit-schema provenance tags carried in the records, the vision-language model together with its digest and quantization, the inference runtime and driver, the text-processing module version, the resolver and memory parameters, and the content hashes of the semantic-object ontology and of the route graph. The working tree was clean at every one of the 214 recorded start-ups, so no decision in the campaign was taken by a build that differs from the committed source.

Three of these fields carry the parameters that govern selective resolution and memory promotion, and they were fixed for the entire

**Table 4**
Campaign provenance and Session E design. Part A separates development-stage Sessions A–C from frozen confirmatory Session E. Part B gives block purposes, experimental units, raw counts and primary groupings. No single global Session E N is reported; corpus totals are not a pooled experimental N, and repeated cycles are not independent semantic-intent cases. All blocks cover B/C except E7, which uses B only.

Part A - campaign boundary

| Campaign | Role | Resolver state | Commit / clean-tree provenance | Audit sessions / decisions | Scenario-classified |
|---|---|---|---|---|---|
| **A** | development | v4.7.4 → v4.8; Step-0 θ = 0.60 | not recorded | 10 / 85 | 37 |
| **B** | development | v4.8; Step-0 θ = 0.60 | not recorded | 3 / 76 | 41 |
| **C** | development | v4.8; Step-0 θ = 0.60 | not recorded | 4 / 4 | 4 |
| **E** | **confirmatory** | frozen v4.8-review (Table 3) | commit and clean tree, all 214 startups | 214 / 226 | not applicable |

Session A is version-mixed. Its 85 decisions split 63 under v4.7.4 and 22 under v4.8, and its 37 scenario-classified decisions split 26 and 11. Sessions A–C are development-stage evidence and are not estimates of final-system performance; no Session E parameter is projected back onto them.

Part B - Session E blocks

| Block | Purpose / component | Experimental unit | Raw observations | Primary N / grouping |
|---|---|---|---|---|
| **E0** | Start gate: digest, graph and ontology verification. *Provenance gate* | Configuration-check session | 11 audit sessions; 10 decision cycles; 2 intents | **Excluded from all primary performance rates** |
| **E1 / E1b** | Post-arrival visual confirmation, presence and absence. *L3 confirmation; static ontology* | Controlled scene-condition trial | 39 decision cycles; 19 intents | **8 E1b trials**: 4 present + 4 absent; the other 31 are descriptive |
| **E2** | Negation precedence, exclusion, location-referent conflict, matched controls. *Negation and location-conflict guards* | Semantic-intent cluster | 49 decision cycles; 19 intents; **39 repeated guard activations across 14 intents** | **19 intent clusters: 14 negation/exclusion intents producing 39 repeated guard activations; 2 location-conflict intents; 3 matched controls.** |
| **E3** | Resolver repeatability, unit declared in advance. *L3a cascade; M3 Step 0* | Semantic intent | 56 decision cycles; 11 intents | **11 semantic-intent clusters**; repetitions retained within cluster |
| **E4** | Rejection of unstable repeated associations. *M3 promotion gate* | Candidate instruction–node association | 23 decision cycles; 4 intents | **2 promotion candidates, each induced from 4 controlled E4.1 observations (8 induction cycles total); the remaining E4.0/E4.3 cycles provide baseline/control/post-compile checks.** |
| **E4b** | Induction → digest compile and install → Step-0 reuse on the other platform. *Promotion; shared-memory reuse* | Candidate association; recipient reuse cycle | 10 decision cycles; 2 intents | **1 induced association**; the 4 recipient reuse cycles verify it, not four learned ones |
| **E5** | Simultaneous two-robot access to the shared L3b inference server. | Concurrent pair | 6 missions; 3 semantic-intent IDs | **3 concurrent pairs** |
| **E6** | Removal of the compiled semantic-memory layer. *M3 Step 0 + M1/M2 prompt prefix, jointly* | Instruction × platform paired unit | 4 memory-off decision cycles; 2 instructions | **4 paired units** against memory-on observations |
| **E7** | Sealed unseen-instruction evaluation. *Full L3 pipeline* | Sealed instruction cell | 29 decision cycles / 28 sealed cells | **28 sealed cells**: **24 in the P1–P4 grid** (six families × four levels) **plus 4 G cells**; the duplicate cycle is descriptive |
| **Totals** | | | **214 audit sessions; 226 decision cycles** | **No pooled primary N** |

Session E contains 214 audit sessions, of which 210 are decision-bearing, yielding 226 decision cycles; four audit sessions contain no decision cycle (E0 × 1, E2 × 2, E3 × 1). These are corpus totals, not a pooled experimental N.

campaign rather than tuned during it. The Step-0 acceptance threshold is 0.75; a candidate is accepted only if it also leads the runner-up by at least the 0.10 abstention margin, and the resolver abstains otherwise. An instruction–node association is promoted into compiled memory only when it has been observed at least three times, its rounded consistency is at least 0.80, and at least one accepted L3b resolution is among the observations. These values are stated here in their normative form because they define what the deterministic path was permitted to do; Sections 4.4 and 5.3 describe the mechanisms they parameterize, and Section 7 reports what the system did under them. The sensitivity analyses reported later are post-hoc robustness characterizations and were not used to optimize these frozen values.

The provenance record contains one ambiguity that we report rather than resolve. All 214 start-up records carry two schema fields with different values, audit_schema = 2.2 and audit_schema_version = 2.1. The frozen configuration record fixes audit_schema_version = 2.1 as the provenance identity, and Table 3 names the field it reports for that reason: a bare schema version would therefore be ambiguous in the raw record, whereas treating the two fields as equivalent would assert a correspondence that the records do not document. Both are therefore retained, under their own names. The ontology and route-graph digests are given for the same purpose: they identify the semantic map and the navigation graph the campaign ran against, and the route-graph digest was identical when read from either context server, confirming that both robot configurations were exposed to the same canonical route-graph representation during the campaign.

### 6.4. Session E experimental design

Session E comprises capability-specific blocks with experimental units and primary groupings fixed in the frozen design register (Table 4, Part B). Its 214 audit sessions and 226 decision cycles are corpus totals, not a pooled experimental N: semantic-intent clusters, paired memory observations, concurrent mission pairs and sealed instruction cells are distinct units. Repeated cycles remain grouped within the relevant unit; block-specific populations and denominators are defined in Section 6.5.

One block is excluded from the primary Session E performance rates by design. The start/configuration gate verifies the expected compiled digest, route graph, and semantic-object ontology within its designated configuration-check sessions. It has no performance estimand; its decision cycles are retained in corpus accounting. Primary Session E performance rates exclude E0, whereas the descriptive M1 positive-confirmation prevalence includes the E0 start-gate checks and is reported over all attempted confirmation checks.

The blocks address selective resolution and safeguards, resolver repeatability, memory promotion and cross-platform reuse, compiled-memory removal, post-arrival confirmation, concurrent L3b access and sealed-instruction evaluation. Table 4, Part B gives their conditions, experimental units and groupings; Sections 7.1–7.5 report the corresponding results.

Unlike development Session C, which exercised concurrent L3a missions, the frozen concurrency block tests simultaneous access to the shared L3b inference service.

The sealed block deserves one clarification, because its counts are easy to conflate. It comprises 28 sealed instruction cells in total: 24 of them form a difficulty grid of six instruction families at four levels, and the remaining 4 are cells outside that grid. The grid and the additional cells are reported separately in Section 7.1 for that reason, and the 28 cells are the unit throughout; H-P4 produced two decision cycles within the same run, with the same selected node, and only the first observation per cell is scored.

The first author constructed the E7 instructions and acceptable-destination sets before execution, using the intended referents, knowledge of the site, the route graph and semantic-object information. The original set was not independently semantically validated before execution. Fifteen instructions were first written in Romanian before ontology terms were introduced during test construction; thirteen were written directly in English after such terms had been discussed. The executed English instructions and their keys were sealed before the first mission, and the operator copied them verbatim. Pre-execution sealing establishes that the key was not adjusted to the system's results; it does not establish independent instruction generation or agreement among independent semantic judges.

Acceptable sets encoded the author's intended referent and the admissible mapped destinations, allowing several nodes when the task could be served at more than one location. Clarification counted as correct only for cells whose sealed key required it, G-3 and H-P2; an explicit clarification request was necessary, and a model/parser failure never counted as correct clarification. The W-P2 key encoded a plant referent and excluded the north-window node 14 despite the wording “a plant or sunlight”. The P-P3 key encoded the author's fresh-air interpretation of “Where can I refresh?” at node 14. These author-intent judgments were prespecified, but the natural wording permits broader interpretations; their semantic appropriateness was therefore checked separately.

One independent evaluator, Andy Dumitrescu, a master's student at the authors' faculty who had not constructed or executed E7 or Session E, judged all 28 cells in a post-hoc sensitivity analysis. He received the verbatim English requests in randomized order under neutral codes, together with the route graph, static-object information and a map. The original key, intended-referent annotations, system choices, outcomes and scores were withheld. He listed acceptable nodes or requested clarification and completed an independence declaration. The analysis plan and packet were sealed before dispatch. His judgments were used only as a separate reference; neither the original key nor its primary score was replaced.

All blocks ran on both robot configurations except E7 (Table 4), which assessed semantic destination selection and clarification decisions on Xplorer-B only and provides no cross-platform estimate.

### 6.5. *Outcomes, experimental units, and statistical analysis*

Every reported quantity is tied to an explicitly defined population and analysis unit. At corpus level, Session E contains 214 audit sessions, 210 decision-bearing runs, 226 decision cycles, and 90 distinct semantic-intent identifiers. These counts describe different levels of the audit corpus and are not interchangeable experimental sample sizes. An audit session corresponds to one navigator startup and provides the unit for configuration and provenance. A decision-bearing run is an audit session containing at least one decision cycle and is the unit used for physical navigation outcome. A decision cycle is one recorded semantic decision and is the observation level for resolution path, guard activation, and inference calls. For experimental outcomes, however, the primary unit remains block-specific, as defined in Table 4: controlled scene-condition trial, semantic-intent unit or cluster, candidate association or recipient reuse cycle, concurrent pair, instruction-by-platform paired unit, or sealed instruction cell.

The population and exclusion rule are stated together with each reported outcome. The start/configuration gate contributes to corpus and provenance accounting. The E0 exclusion and descriptive M1 exception are defined in Section 6.4. Physical navigation outcome is evaluated at run level after the start-gate exclusion. The terminal record defines the complete outcome distribution. Separately, all decision cycles of each run are checked for an actual navigation dispatch: terminal-miss incidence uses all protocol-valid runs, whereas terminal misses and non-arrival conditional on dispatch use only runs that sent a navigation goal. Intermediate recovery records within a subsequently completed run are not counted as independent mission failures. Resolution-path composition, guard activation, and inference-call quantities are decision-cycle properties over their stated populations. Block-specific semantic outcomes, including sealed-cell acceptability, candidate promotion or rejection, and recipient-platform shared-memory reuse, are evaluated on the primary unit defined for the corresponding block. Repeated cycles within one such unit are retained as repeatability observations rather than promoted to independent cases. For the sealed-instruction evaluation, acceptability is defined by the predefined acceptable-node set associated with each cell rather than by equality to a forced single-node target; where the protocol permits multiple semantically admissible destinations, any node in that set is scored as acceptable. Escalation to L3b is an intermediate routing decision, not a final outcome. Final semantic outcomes distinguish acceptable selections, incorrect executable decisions, explicit clarification requests and model/parser failures. For E7, incorrect executable decisions are further separated into wrong destinations and executable decisions where the sealed key required clarification. Absence of a node alone does not establish deliberate clarification. The first observation per sealed cell is scored; H-P4 contributes two decision cycles in one run, not two independent trials.

Repeated-measure uncertainty is handled at the level at which independence is defensible. For the architecture-level latency contrast, confidence intervals are obtained from 10,000 bootstrap resamples with fixed seed 20260822, resampling semantic_intent_id clusters rather than individual decision cycles. Resampling individual cycles would treat repeated executions of one intent as independent observations and would therefore understate uncertainty. Other quantities use the uncertainty treatment appropriate to their declared unit; when the design or sample size does not support a meaningful interval, no interval is reported.

The historical Session E latency analyses report three distinct estimands. The architecture-level contrast is an unpaired comparison between naturally selected deterministic-path decisions and accepted L3b decisions after exclusion of the start gate; the arms contain different instructions and therefore do not constitute a causal VLM-only baseline. The latency ratio itself is used as the interpretable effect-size estimand, with uncertainty obtained by cluster bootstrap over semantic intents; the submitted decision-cycle Mann–Whitney comparison is not retained because it would treat repeated observations within an intent as independent. The second latency estimand is the paired within-instruction, within-platform contrast from the compiled-memory removal experiment; with four paired units, it is reported as a median and range rather than with a sampling interval. The third is the within-case before/after contrast for the single promoted association subsequently reused on the recipient platform; because its pre-promotion baseline contains one inference call, it is reported only as a mechanistic illustration and not as an estimate with sampling precision.

The controlled comparison in Sections 6.8 and 7.7 separately measures offline instruction-to-decision latency and resolver-compute latency on paired instructions and contexts. It is distinct from the historical contrasts above.

The external localization reference is analyzed separately from these Session E performance estimands. Its observation unit is an included tag detection; 11 detections satisfy the inclusion criterion, with 10 originating from one tag and one from the opposite-normal tag. Its observed dispersion and the single-tag-dominance limitation are reported separately in Section 6.6.

### 6.6. External localization reference

The external localization reference is an opportunistic cross-block analysis rather than an additional Session E block. No dedicated run was performed for this analysis, and no part of the navigation stack reads the markers: five AprilTag markers of one family and a common 125 mm side are fixed to the walls of the test floor, installed for an unrelated docking experiment, and three of them have surveyed map coordinates. Whenever a mission image happens to contain one, the four tag corners are expressed directly in map coordinates from the surveyed center and wall normal and solved for the camera pose with the calibration carried on the robots; the base-link estimate that follows can then be compared with the AMCL pose the robot acted on in the same record. Working directly in map coordinates removes any tag-local frame, and with it any convention about the direction in which a tag's own axes point.

The observation unit is one included detection. Detection was run at native resolution and twice that resolution, retaining the larger detected image-scale observation and rescaling its corners. Across 345 mission images, 29 detections were obtained. Two were excluded because the marker was not surveyed, twelve because the apparent mean tag side was below the prespecified 80-pixel threshold, and four because their detections were inconsistent with the surveyed marker pose. No pose-solver failure was recorded. This leaves 11 included detections. The size threshold was fixed before analysis; the survey-consistency exclusion was applied after inspecting the data and remains a declared post-hoc exclusion.

For each discrepancy component we report the included count, sample mean, sample standard deviation using $n - 1$ in the variance denominator, and observed minimum and maximum. These summaries describe the observed detections. We do not report an interval estimate for the mean or treat the detections as independent measurements: ten originate from tag 3 and one from tag 4, with seven observations on Xplorer-B and four on Xplorer-C.

Three limitations bound this reference. The opposite-wall-normal comparison rests on a single detection, so it does not establish a map-wide registration pattern or identify the source of the discrepancy. Sampling is opportunistic and depends on the mission images, marker placement and camera framing; repeated detections from one marker share systematic influences. Finally, the reference was collected during Session E, after the development-stage Sessions A–C. It therefore supplies a contemporaneous external check for Session E at this site, but cannot retrospectively ground-truth an individual A–C navigation outcome or establish a cause for a navigation miss.

### 6.7. Controlled L3b model-substitution replay

The controlled model-substitution replay is a post-hoc offline analysis of archived Session E records, not an additional experimental block and not a second physical campaign. Its question is narrow: whether the frozen L3b application interface - the prompt, the parser and the context contract exactly as the confirmatory system used them - can be driven by vision-language models other than the one Session E ran on, without any model-specific change to that interface. Nothing in the interface was adapted for the substitutes. The control arm is the Session E model and digest itself. All arms used the same replay harness, the same frozen prompt, parser and context contract, the same controlled sampler settings, and the same per-record seed schedule; model identity was the intended arm-level substitution.

The cohort is defined by a funnel rather than by a sample size, because the reduction at each step is what determines how the results may be read. Session E contains 105 attempts on the visual-language path; excluding the start gate leaves 104. Of these, 22 are records in which the model returned no valid destination, yielding a clarification outcome, and 14 are records in which no accepted result was produced at all. Neither group has a persisted pre-navigation image, which is the input a multimodal counterfactual requires. For the 22 clarification records the parsed proposal of no valid destination is preserved, whereas the raw reply text and reason are not; the 14 error or parse records likewise lack the persisted image required for multimodal replay. That leaves 68 replayable records after the start-gate exclusion, and 69 including it, each replayed through three models at three seed bases.

That funnel makes the cohort outcome-conditioned, and the consequence is stated rather than left to be noticed. The 68 records are exactly those on which the original system reached an accepted destination with a complete image, so within the cohort the archive never abstains. This is a property of how the cohort was built, not of the model that built it, and every agreement figure inherits it. For the same reason the original system's own clarification behavior, measured over the full population of guard activations, is descriptive context and not a denominator-matched comparator for the replay abstention rates. Agreement with the archived destination is likewise reference agreement against one recorded stochastic outcome, not semantic accuracy: the archive is an outcome, not ground truth.

Two analysis conventions follow from the design. The 68 records are the clustering units; the three seed outputs produced for each record and model are repeated observations within that record and are not treated as independent experimental units. Three deterministic seed bases were used to characterize decoder variability. For each base, the requested seed was derived deterministically from the record key and applied identically across model arms. This schedule controls the requested random-number initialization, but it neither guarantees bit-identical inference on the accelerator nor establishes statistical independence between records. Latency is reported descriptively on the common replay workstation, using both absolute wall-time summaries and ratios to the control arm; these quantities are not interpreted as hardware-independent model benchmarks.

The scope of what this replay can support is therefore bounded in one direction only. It bears on portability of the L3b application interface across the evaluated models: whether an alternative model can consume the unmodified multimodal prompt and return the required destination contract. It does not support semantic model invariance, which is not claimed anywhere in this paper, and it does not estimate alternative-model behavior over all 104 attempts on the visual-language path, because it never saw the ones the original system did not resolve. The replay also covers only the L3b semantic destination-selection path; the separate post-arrival visual-confirmation interface, which uses a distinct prompt, endpoint and inference configuration, was not included. Section 7.6 and Table 5 report the outcome under exactly these limits.

### 6.8. Controlled comparison of semantic resolution and offline decision latency

Three arms resolved the same instruction-context items: A, the complete system (M3 Step 0, deterministic Steps 1–6, VLM on escalation); B, Steps 1–6 with M3 disabled; and C, direct VLM resolution bypassing Steps 0–6. All retained the same item context, M1/M2 prefix, forbidden-node envelope, VLM prompt, response contract and post-response guard/structural validation. A–B isolates M3 target reuse conditional on this shared safety input; B–C compares deterministic selection with direct VLM resolution; A–C compares complete routing configurations. C is therefore a VLM-only target resolver under common safeguards.

The standardized condition emptied the dynamic-object store while retaining each item's archived start image, pose, static objects, graph, memory and safety envelope. It contained 38 items: five familiar associations, 27 unseen formulations and six clarification-required cases. Nine image-incomplete items were excluded without substitution: E7 S-P2, P-P4 and H-P3, plus six E2 instructions listed in the exclusion manifest. Archive missingness is not assumed random. E7 parents were the first frozen observation per cell; familiar and E2 parents were the first chronological protocol-valid observation with a start image, not selected for success. For E2-X02/X03 this was repetition two: the first

**Table 5**
Controlled L3b model-substitution replay through the unchanged prompt/parser. Full replay: 69 records × 3 models × 3 seeds = 621 outputs; primary E0-excluded analysis: 68 × 3 × 3 = 612 (204 per model). Records, not seed outputs, are the clustering units. The cohort contains only archived accepted-node decisions with retained images, excluding unresolved/image-incomplete attempts from the 104 E0-excluded VLM-path attempts (105 including E0). Archived-node agreement is reference agreement, not semantic accuracy. Interface portability was observed across the tested models without prompt/parser adaptation; semantic behavior and low-sample promotion remained model/sampling dependent.

Part A - per-model replay behavior (204 outcomes per model, over 68 records)

| Model | Parser compliance | Accepted node | Abstention (node_id = −1) | Archived-node reference agreement | Same node in all 3 seeds | Median latency (ratio to control) |
|---|---|---|---|---|---|---|
| **qwen3.5:4b** (control) | 181/204 = 88.7% | 160/204 = 78.4% | 21/204 = 10.3% | 137/204 = 67.2% | 40/68 = 58.8% | 7.20 s (reference) |
| **gemma4:e2b** | 204/204 = 100.0% | 194/204 = 95.1% | 10/204 = 4.9% | 131/204 = 64.2% | 66/68 = 97.1% | 6.77 s (0.94 ×, IQR 0.89–0.97) |
| **qwen3.5:9b** | 199/204 = 97.5% | 165/204 = 80.9% | 34/204 = 16.7% | 120/204 = 58.8% | 50/68 = 73.5% | 14.63 s (2.01 ×, IQR 1.87–2.18) |

**The control uses the Session E model/digest with the common replay sampler. Its agreement measures same-model replay against the archived run under seed substitution; the archive is not one of the three replay seeds.**

**Seed unanimity uses all 68 records, including parsing failures as non-unanimous. Among records with three parsed responses (48, 68, 63 in column order), unanimity is 40/48, 66/68 and 50/63; this conditional view separates parsing from destination variability.**

**Contract failures were 23, 0 and 5 in column order. The control's 23 formatting failures affected 20 records: 13 backslash-escaping, nine truncation and one other; 14 contained a recoverable integer. None of the 621 calls had transport failure or retry.**

**Negation/exclusion subset: 13 records and 117 outputs; no forbidden proposal occurred. Neither replay nor original observations required the post-VLM guard to block a proposal, so this is defense-in-depth rather than an observed blocking test.**

Part B - M3 promotion stability across the nine model × seed arms

| Candidate association | Archived gate decision | Arms promoting | Reading |
|---|---|---|---|
| **quiet_corner** | **reject** - frequency 4, dominant node 9, consistency 0.75 | **8 / 9** | the archived rejection is **not robust**; the one non-promoting arm is qwen3.5:9b at seed 3 |
| **wet_gloves** | **promote** - frequency 4, consistency 1.00 | **9 / 9** | robust under model and seed substitution |
| **deliveries** | **reject** - frequency 4, consistency 0.50 | **0 / 9** | rejection is stable, but this is a **partial** counterfactual only: one of the four original images is missing |

Each candidate had four observations and met the accepted-L3b condition, leaving consistency $\geq$ 0.80 to determine promotion. The gate measures sampled-output stability, not semantic truth; quiet_corner illustrates a model/sampler-dependent low-sample decision.

lacked an image and had an adverse outcome. “Unseen” means not stored in the applicable memory, not independent investigator-unseen generation.

The historical condition restored the same-cycle archived dynamic prompt context for 28 items (23 unseen, five clarification-required), holding other inputs constant. It reconstructs the printed context and distance constraints, not exact historical perception geometry, and contains no familiar-association stratum. Conditions were analyzed separately; context sensitivity used only their 28 common items. Familiar items were scored against the parent's stored historical association, E7 against the original sealed acceptable sets, and E2 against fixed prohibition/exclusion references. Scoring rules were fixed before execution and withheld from target resolution.

Each item used three paired model seeds and a fixed schedule balancing arms across all three within-item positions. The instruction-context item was the primary unit; seeds were repeated observations, with majority correctness defined as at least two correct seeds. Outcomes distinguished acceptable selection, incorrect executable decision, explicit clarification and model/parser failure. An allowed target outside the reference, including execution when clarification was required, was incorrect; parsing failures were always incorrect and remained separate from clarification. Paired summaries counted items better/worse/tied in correct-seed count, wrong executable observations and affected items, and VLM calls. A stratified paired item bootstrap (10,000 resamples) preserved all arms and seeds within sampled items for descriptive correctness and timing uncertainty.

Both conditions used Q4_K_M qwen3.5:4b with fixed settings on an RTX 5050 Laptop GPU and Ollama 0.30.2. Every observation started a dedicated server and fresh runner, preloading the model with an empty request outside the decision timer. A twelve-call qualification preceded each condition and was excluded from results. Process identities, empty pre-load inventory, model digest/context length, GPU placement and unchanged runner verified a resident model processing its first item prompt/image without prior item-prompt reuse. VLM payload hashes matched across arms for the same item, condition and seed; response equality was checked, not enforced.

A monotonic timer measured offline instruction-to-decision latency from raw instruction processing through the executed resolver, archived-image reading/encoding when needed, prompt/payload construction, local API transport/inference, decoding, parsing and validation. Context/index preparation, qualification, model startup/preload, scoring and file writes were excluded. Resolver computation was timed separately for A/B, including escalations, and was inapplicable to C. All outcomes, including failures, remained in timing summaries. Per-item three-seed medians supplied the item-level paired differences and ratios. Live camera acquisition, robot communication and physical navigation were outside this measurement.

### *6.9. Matcher decision boundaries and offline guard test*

We evaluated the frozen M3 matcher offline on a labeled diagnostic set of 47 instructions: 16 valid paraphrases of stored associations, 15 instructions with high lexical overlap but different meanings, and 16 instructions with competing destinations near the runner-up margin. No robot or VLM was used, and these results were not pooled with the controlled semantic comparison. The set was constructed by the first author with access to similarity scores; each label was fixed before the candidate was scored and was reviewed without scores or matcher decisions, and the final set of 23 destination and 24 abstention labels was sealed before the matcher comparison.

The first two families used the historical memory index of six stored examples mapping to five nodes. Exact-fraction Jaccard scores placed candidates at 1.0, within (0.75, 1.0), at 0.75, or within [0.50, 0.75). Candidates filled these bands in authorship order, with a deterministic coverage rule ensuring representation of all six stored examples; matcher decisions and agreement with labels were not used for selection. One high-overlap, different-meaning item at Jaccard 1.0 could not be supplied naturally, leaving that family at 15 items. All items lacked negation, and none exactly matched a stored example after normalization.

The historical index could not activate the 0.10 runner-up margin when the top score met 0.75. The competing-destination family therefore used the historical index plus eight explicitly synthetic pairs of associations, each mapping to two distinct real graph nodes. This additional index was hashed before candidate scoring. All 16 items had a top score of at least 0.75; six had an exact gap below 0.10, four had a gap of exactly 0.10, and six had a gap in (0.10, 0.25]. Ranking used the maximum score per node across the combined index. Two items involving a semantically questionable synthetic destination association were retained and flagged in the released records.

The published matcher reimplementation was evaluated unchanged at Jaccard 0.75 with margin 0.10, preserving its floating-point arithmetic. An exact-match memory baseline accepted only a unique destination associated with an identical normalized instruction; otherwise it abstained. Jaccard 0.60 with the same margin was retained as a separate historical reference. Outcomes distinguished correct acceptance, wrong acceptance, abstention when a destination was expected, and correct abstention.

A separate deterministic branch test exercised the frozen deployed navigator, including its parser, post-response negation guard, action validation and fallback. It used 14 distinct E2 negation instructions with graph, pose, memory and static semantics derived from their parent records and an empty dynamic-object store. Synthetic replies were injected at HTTP level and navigation dispatch was intercepted. The 139 cases included every forbidden integer proposal, permitted-node controls, explicit clarification, and malformed, missing or incorrectly typed targets. Expected branches were specified from the frozen code. The same deterministic cases had been exercised for debugging before sealing, so this was an implementation conformance test with known outcomes, not a blind evaluation. The recorded run completed all cases.

## 7. Results

### *7.1. Deterministic reasoning, guard behavior, and sealed-instruction outcomes*

After exclusion of the start-gate block, the Session E corpus contains 216 decision cycles. Of these, 112 were resolved by the deterministic cascade without a model call and 104 entered the L3b path. Fig. 5(a) shows the composition. This proportion describes the instruction mix actually exercised in this campaign and is not a resolver-wide rate: the blocks were built to interrogate different capabilities, and their fast-path shares differ by design, from every cycle in the repeatability block to none at all in the promotion and concurrency blocks.

Within those 112 deterministic resolutions, the first rule that resolved the instruction was the learned-association step in 39 cases, an explicit node identifier in 18, a node name in 17, an object identifier in 8, an attribute match in 13, a single class match in 3, and class-with-proximity in 14. The logged deterministic resolver-compute latency had a median of 0.245 ms and a maximum of 0.960 ms, and all 112 observations were below 1 ms; the deterministic path is therefore reported as sub-millisecond, at that resolution rather than as a smaller round figure. This quantity is the resolver's own computation time, not an end-to-end latency for the deterministic path: the pre-navigation image is acquired before the timer starts. These counts record which step resolved each instruction first. They do not estimate what removing a step would cost, because an instruction unresolved at one step may still be resolved by a later one or escalate to L3b.

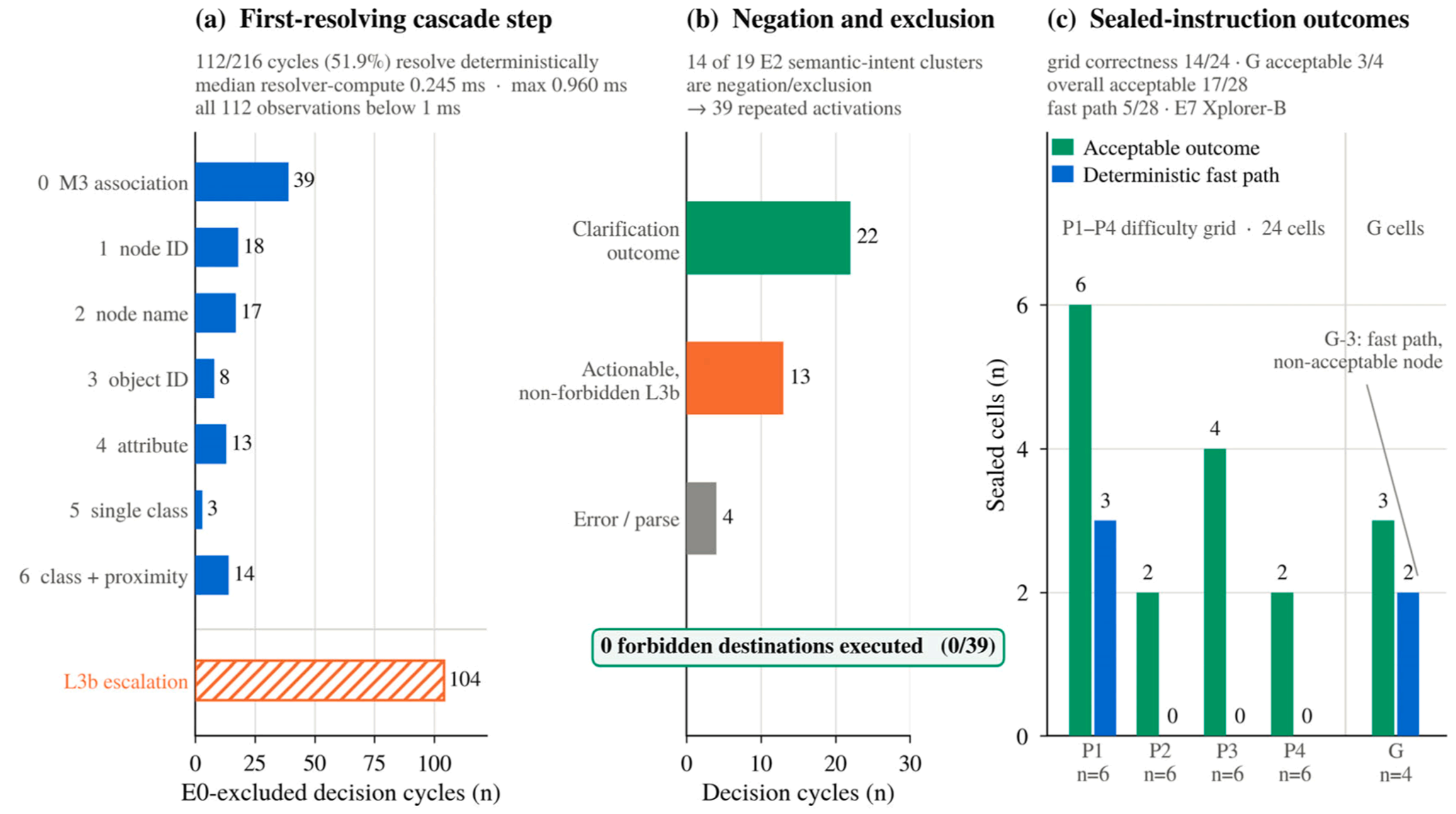

**Fig. 5.** Deterministic reasoning, guard behavior, and sealed-instruction outcomes in Session E. (a) First-resolving step over 216 E0-excluded cycles: 112 (51.9%) were deterministic (median resolver-compute 0.245 ms; maximum 0.960 ms; all below 1 ms), and 104 escalated to L3b (hatched). (b) Fourteen of 19 E2 semantic-intent clusters produced 39 repeated negation/exclusion activations: 22 clarification outcomes, 13 actionable non-forbidden resolutions and four parse/escalation errors; no forbidden destination was executed. (c) E7 on Xplorer-B: 28 sealed cells scored on their first observation (29 records; H-P4 produced two cycles in one run). Acceptable selections were 14/24 in the P1–P4 grid and 3/4 in G, totaling 17/28. Complete outcomes were 17 acceptable selections, six wrong destinations, two executable decisions where clarification was required, three parsing failures and zero clarification requests. Five cells used the deterministic fast path; incorrect G-3 gives false-fast incidence 1/28 and conditional error proportion 1/5. Independent-reference sensitivity is reported separately in Section 7.1.

For the architecture-level contrast defined in Section 6.5, the ratio of the median decision latency among accepted L3b decisions to the median decision latency among deterministic-path decisions was 45,443 ×, with a 95% semantic-intent cluster-bootstrap interval of 30,428–70,512 ×, using 40 semantic-intent clusters in each arm. These analysis sets are specific to the latency estimand and do not constitute a partition of the 90 semantic-intent identifiers in the full Session E corpus. This is a contrast between naturally selected execution paths, not a causal VLM-only baseline or a per-instruction performance guarantee.

The negation and location-conflict block contains 19 semantic-intent clusters: 14 negation/exclusion intents, two location-conflict intents, and three matched controls. The 14 negation/exclusion clusters produced 39 repeated guard activations; these are repeated observations within 14 clusters, not 39 independent semantic cases. None of the 39 activations was served by M3 at Step 0. Their outcomes comprised 22 clarification requests, 13 actionable resolutions to a node not forbidden by the instruction, and 4 parse or escalation errors. Thirteen of the 39 therefore produced an executable destination, and no forbidden destination was executed. The two location-conflict intents were detected and escalated. For the negation and exclusion subset, bypassing ordinary deterministic pattern matching is intentional: such instructions are routed to constrained escalation or to a clarification request rather than accepted through a potentially conflicting match. The post-response safeguard is consequently interpreted as defense in depth: among the activations that reached it, the model did not propose a forbidden destination, so no observed case required that safeguard to block one.

The repeatability block comprises 11 semantic-intent clusters over 56 decision cycles, every one of which resolved deterministically with no model call. Destination selection was unanimous within all 11 clusters, and the stricter combination of resolution method, cascade step and destination was likewise invariant within all 11; together, the clusters exercised deterministic Steps 0, 2, 3, 4 and 6. Because the semantic-intent cluster is the experimental unit, the 56 cycles are repeated observations within 11 units. The result therefore supports within-intent resolver-path repeatability under the tested frozen configuration, not semantic correctness.

The sealed-instruction block was executed on Xplorer-B and comprises 28 cells with acceptance criteria sealed before execution. It contains 29 decision records because H-P4 produced two cycles in one run: a navigation miss followed by completion at the same node 9, which was outside the acceptable set. The first observation per cell is scored. Nineteen cells permitted multiple destinations. Correctness was 14/24 in the six-family difficulty grid, with 6/6, 2/6, 4/6 and 2/6 at P1–P4, respectively, and 3/4 in the additional cells, giving 17/28 overall. Five cells resolved deterministically, three in the grid and two outside it.

The corrected final-outcome classification is 17 acceptable selections, six wrong executable destinations, two executable decisions where clarification was expected, three model-response parsing failures and no clarification requests. H-P3, P-P4 and S-P2 escalated to L3b but ended in parsing failure; they were not deliberate abstentions. G-3 and H-P2 received executable destinations despite requiring clarification under the sealed key. Thus neither clarification-required cell elicited clarification, and the three missing-node outcomes reflect parser failure. The corrected taxonomy leaves the primary 17/28 score unchanged.

G-3, “Where is the H sign?”, resolved at Step 5 to node 23 and executed, although its sealed key required clarification for an unmapped referent. The best-supported reconstruction is a partial class match on “sign” to a YOLO stop-sign object attached to node 23. That object appears in the preceding G-2 prompt with confidence 0.70 and in the archived dynamic store; no static stop-sign class was present. Step 5 permits a partial class-name match, while the subsequent checks validate the action, graph membership and distance rather than the intended sign's identity. These checks therefore did not establish semantic correctness. The exact G-3 context was not persisted, so the mechanism remains a reconstruction from frozen code and surrounding records. False-fast incidence was 1/28 across sealed cells (Wilson 95% interval 0.006–0.177), and the error proportion among fast-path decisions was 1/5 (0.036–0.625). Both denominators are small. The frozen resolver and historical results were not modified.

The independent evaluator agreed with the sealed verdict of destination versus clarification in 26/28 cells. Among the 25 cells in which both references specified destinations, 14 node sets were identical, five were evaluator subsets, two were supersets and four partially overlapped. Rescoring under the evaluator's judgments gave 16/28 instead of 17/28. G-1 and L-P3 changed because the evaluator excluded CB204, which serves as both a laboratory and an office; W-P2 changed because he accepted the north-window node. For P-P3, he accepted nodes 0, 4, 14 and 20; the system's node 9 remained incorrect. These judgments form a separate sensitivity analysis, preserving the original sealed score. Fig. 5(c) reports the historical sealed-cell results.

The sensitivity analysis of the learned-association step characterizes the operating point the campaign ran under; it was performed after the campaign and was not used to choose that point. A standalone reimplementation, reading only released inputs, reproduced the accept-or-abstain outcomes for all 212 E0-excluded decisions with non-empty M3 state and all 39 accepted destinations at the frozen threshold and margin. The 39 acceptances arise from 5 distinct instruction strings, so they too are repeated observations rather than independent cases.

Two properties of that operating point follow. All 39 accepted matches scored an exact 1.0, while the highest-scoring rejected candidate scored 0.625, so the matcher returns the same decisions for thresholds strictly above 0.625 and up to 1.0; the frozen threshold sits inside that plateau rather than on an edge. The abstention margin, in turn, is inactive over the whole range from 0 to 0.80 at the frozen threshold, because every accepted multi-candidate match is separated from its runner-up by at least 0.80. Session E therefore does not identify the frozen margin as a uniquely correct value, and it is reported as a conservative ambiguity guard rather than as a tuned quantity. Lowering the threshold to the value used during development would admit four further matcher-level acceptances whose candidate differs from the protocol-defined expectation; these are counterfactual acceptances at the matcher, not demonstrated navigation errors, since the complete resolver applies downstream location-conflict handling that this analysis does not model.

The labeled matcher-boundary and exact-match comparisons are reported separately in Sections 6.9 and 7.8; they do not replace these historical operating-point results.

### *7.2. Memory promotion, ablation, and shared-memory reuse*

Promotion outcomes are analyzed at the level of the candidate instruction–node association. Session E produced three controlled candidates, each observed four times and each containing at least one accepted L3b resolution; all therefore satisfied the frequency and accepted-L3b conditions of the frozen promotion rule. Their nominal decisions consequently depended on destination consistency against the 0.80 threshold. wet_gloves was unanimous at 1.00 and was promoted, whereas quiet_corner and deliveries reached 0.75 and 0.50 and were rejected. The latter two are the rejection candidates of E4. The promotion decisions for the two E4 rejection candidates are based on the eight controlled E4.1 induction cycles; the remaining E4 cycles belong to baseline, control, or post-compile phases and are not additional promotion observations. Fig. 6(a) shows the three outcomes. Promotion is therefore interpreted as a sampled-output stability criterion, not semantic correctness.

E4b demonstrates the complete promotion-to-reuse chain. A baseline cycle on the recipient platform, Xplorer-B, required L3b for the target instruction and independently selected the same destination later induced on Xplorer-C. Four subsequent induction observations on Xplorer-C, under the same pre-promotion memory state, also resolved through L3b and selected the same node. These observations yielded one candidate association, which passed the gate and was compiled and installed in the post-promotion digest. Four subsequent cycles on Xplorer-B then resolved the same instruction deterministically at Step 0, with Jaccard 1.00 and no model call. A final near-miss instruction matched no stored association and returned to L3b. Fig. 6(b) shows the ten cycles in order. The four recipient-platform cycles are repeated verifications of the same single induced association, not four independent associations. The demonstrated mechanism is therefore shared-memory reuse across robot platforms through deliberate digest compilation and installation, not direct robot-to-robot learning.

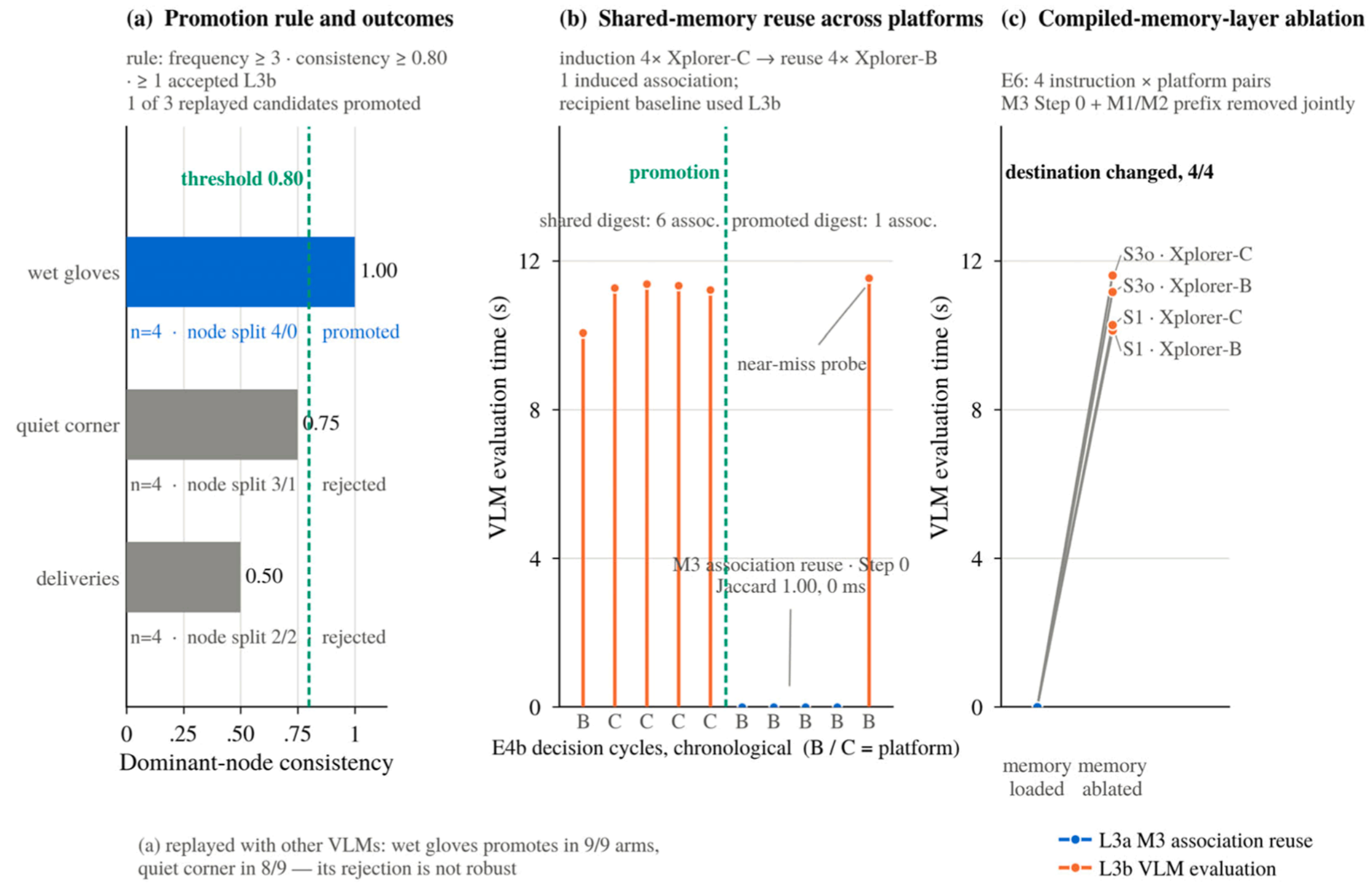

**Fig. 6.** M3 promotion, managed cross-platform reuse and compiled-memory-layer ablation. (a) Post-hoc gate replay for three E4/E4b candidates, each observed four times. Under frequency $\geq 3$, consistency $\geq 0.80$ and at least one accepted L3b resolution, wet_gloves (1.00; 4/0 node split) promotes; quiet_corner (0.75; 3/1) and deliveries (0.50; 2/2) do not. Promotion measures sampled-output stability, not semantic correctness. Model/seed replay promotes wet_gloves in 9/9 arms and quiet_corner in 8/9, exposing instability of the archived rejection. (b) Ten E4b cycles: one Xplorer-B baseline, four Xplorer-C induction cycles under the six-association digest, deliberate installation of a one-association digest, four Xplorer-B Step-0 reuse cycles (Jaccard 1.00; no VLM call), and one near-miss L3b cycle. All reached the radiator node. The four reuse cycles verify one association, not four independently learned mappings. Zero VLM evaluation time denotes no model call, not zero instruction latency. (c) Four E6 instruction-by-platform pairs jointly remove M3 Step 0 and the M1/M2 prefix. All memory-loaded units resolved at Step 0; all ablated units invoked L3b (10.1–11.6 s) and changed destination (S1: toilet_m → radiator; S3o: plant_5 → junction_e1).

E6 probes the compiled memory layer on four instruction-by-platform paired units. For every unit, the memory-loaded baseline observations resolved unanimously at Step 0 without a model call. Under the empty-memory intervention, all four decisions escalated to L3b and all four selected a destination different from their corresponding memory-loaded baseline. The paired memory-off / memory-loaded decision-latency ratio had a median of 75,500 $\times$ (range 72,039–88,269 $\times$ across the four paired units); no sampling interval is reported. Fig. 6(c) summarizes the paired contrast.

The ablation does not isolate M3 from all other memory effects. The empty-memory intervention jointly removes M3 Step-0 matching and the M1/M2 prefix supplied to L3b. The transition from Step 0 to L3b follows from removal of the matching path that had resolved the memory-loaded observations, but the changed L3b destination cannot be attributed to M3 alone because prompt context is removed simultaneously. M1/M2 are therefore only jointly probed here, not independently evaluated.

A separate post-hoc sensitivity analysis characterized the frozen promotion-gate boundary without changing the rule used during Session E. With the consistency threshold fixed at 0.80, minimum-frequency settings from 1 through 4 produced the same outcome - only wet_gloves was promoted - whereas a minimum frequency of 5 or greater rejected all three controlled candidates. Session E therefore does not empirically identify 3 as an optimal minimum-frequency setting, because all three candidates were observed four times. With minimum frequency fixed at 3, consistency thresholds up to 0.50 promoted all three candidates, thresholds from 0.51 to 0.75 promoted quiet_corner and wet_gloves, and thresholds from 0.76 to 1.00 promoted only wet_gloves. The frozen 0.80 consistency threshold therefore lies within the observed 0.76–1.00 plateau. This is a post-hoc robustness characterization of a pre-frozen rule: the sweep was performed after the campaign and was not used to choose either parameter value.

Model substitution provides a separate robustness check: all three candidates were replayed across three models and three seed bases, giving nine arms. Within those arms, wet_gloves promoted in 9/9 arms and quiet_corner in 8/9, whereas deliveries promoted in 0/9; the latter remains a partial counterfactual because one original image is unavailable. The 0.75 candidate therefore shows that a low-sample gate decision close to the threshold can change with model or sampling behavior. This reinforces the interpretation of promotion as a stability gate over observed outputs rather than a semantic-accuracy test.

### 7.3. *Post-arrival visual confirmation*

The controlled E1b experiment separates physical arrival from post-arrival visual-semantic confirmation. It tests the corrected fire-extinguisher referent at CB202, node 8, with an object-removal condition. Eight scene-condition trials were run in total: four with the extinguisher present and four with it removed, with two trials per condition on each robot configuration. All eight reached the selected node. The four present trials returned confirmed=True with HIGH confidence, and the four removed-object trials returned confirmed=False with LOW confidence. Thus the outputs agreed with the imposed scene condition in 8/8 trials, while positive-confirmation prevalence was 4/8. E1b does not directly test the earlier situation in which a wall-mounted target was present but outside the camera's field of view.

M1 statistics are recomputed from recorded entity visits during memory extraction and digest compilation. Arrival and confirmation are counted separately. Arrival prevalence is the number of visits ending in an arrival outcome divided by all recorded visits to the entity's node; the accepted arrival outcomes are mission_complete, mission_complete_fallback and mission_complete_same_node. Confirmation prevalence is the number classified as confirmed divided by visits minus not_attempted and unobservable. When this denominator is zero, confirmation prevalence is undefined rather than zero.

A successful check increments the confirmed count, including a positive pose-based check; pose confirmation supports arrival at a location, not visual detection of an object. An attempted check that does not confirm an observable referent is classified as not_confirmed, including correct non-confirmation after object removal, and remains in the confirmation denominator. If all mapped expected objects are marked unobservable, an unsuccessful check is classified as unobservable and excluded from that denominator. A record without a confirmation attempt is not_attempted and is also excluded. Label mismatch is a separate unsuccessful-check status and remains in the denominator. These status rules therefore measure positive-confirmation prevalence, not whether every confirmation or refusal was semantically correct.

Across all 226 Session E decision records, including E0, the status counts were 153 confirmed, 12 not_confirmed, 61 not_attempted, zero unobservable and zero label_mismatch. Across all attempted confirmation checks, including the E0 start-gate checks, 153/165 produced a positive confirmation (92.7%): 23/23 pose-based, 34/34 VLM-contextual and 96/108 VLM-landmark. Primary Session E performance rates exclude E0, whereas the descriptive M1 positive-confirmation prevalence includes the E0 start-gate checks and is reported over all attempted confirmation checks. Accordingly, the reported M1 confirmation statistic is 153/165 over all attempted confirmation checks, including E0; the E0-excluded value is 146/156 and is not used as the primary descriptive confirmation statistic. This is a descriptive property of the recorded checks under the available signatures and observability conditions, not a general visual-verification accuracy estimate. Navigation arrival, positive confirmation and semantic task success remain distinct outcomes.

### 7.4. Concurrent L3b operation

E5 tests simultaneous access by the two robot configurations to the shared L3b inference service. The experimental unit is the concurrent pair: three pairs were executed, comprising six missions in total, one request per platform in each pair. All six missions completed and all six model responses parsed successfully. The block therefore demonstrates functional two-robot use of the shared inference service under the tested concurrency condition.

Inference was slower under this small concurrent workload. The six E5 calls had a mean VLM evaluation time of 13.809 s, compared with 9.785 s over the 84 timed non-E5, E0-excluded VLM calls used as the descriptive non-concurrent comparator. Within each concurrent pair, the absolute difference between the two VLM evaluation times ranged from approximately 6.8 to 8.1 s; the slower platform was not consistent across the three pairs. The comparison is observational rather than a randomized concurrency benchmark, and the available instrumentation does not identify the mechanism producing the additional delay. It therefore supports concurrent-operation feasibility with an observed latency cost, not fleet-scale scalability or a claim of measured server serialization.

### 7.5. Physical execution and external localization reference

Semantic target selection and physical navigation are analyzed separately. After exclusion of the E0 start gate, 200 decision-bearing runs were available; one was invalidated during the campaign and repeated under protocol, leaving 199 protocol-valid runs. Their complete terminal distribution is 162 mission_complete, one missed, 22 clarification_requested and 14 vlm_failed. On Xplorer-B, the 106 runs comprise 87 completed, zero terminal misses, nine clarification requests and ten model/parser failures; on Xplorer-C, the 93 runs comprise 75 completed, one terminal miss, thirteen clarification requests and four model/parser failures.

Checking all cycles within each run identifies 164 runs with a dispatched navigation action, 88 on Xplorer-B and 76 on Xplorer-C. Terminal-miss incidence over all valid runs is therefore 1/199 (0.50%), while terminal-miss incidence conditional on navigation dispatch is 1/164 (0.61%). Two dispatched runs did not end in arrival, giving conditional non-arrival of 2/164 (1.22%). The additional non-arrival is an Xplorer-B run that missed its goal and subsequently ended in model-response parsing failure. It is included in the 14 terminal vlm_failed outcomes, which explains why the dispatch denominator is 164 rather than the 163 runs ending in completion or terminal miss. Wilson 95% intervals, expressed as proportions, are 0.0009–0.0279 for 1/199, 0.0011–0.0337 for 1/164 and 0.0034–0.0434 for 2/164.

The same distinction applies to the platform comparison. Terminal misses over all valid runs are 0/106 on Xplorer-B and 1/93 on Xplorer-C; conditional on dispatch, they are 0/88 and 1/76. Conditional non-arrival is 1/88 and 1/76, respectively. With these few events, the comparisons are descriptive and do not establish a platform difference or equivalence. Mean recorded endpoint error among the completed runs was 0.225 m on Xplorer-B ($n = 87$) and 0.170 m on Xplorer-C ($n = 75$). Eighteen decision records carry missed, including the protocol-invalidated run and recovery cycles; they are not eighteen independent terminal failures. Neither the incidence over all runs nor the conditional navigation measure establishes end-to-end semantic task success: physical arrival can follow an incorrect semantic selection.

The platform differences listed in Table 3 were not experimentally isolated, so navigation outcomes cannot be attributed uniquely to the semantic stack or navigation algorithm. Fig. 4 provides the physical and environmental context.

The separate AprilTag reference contains 11 included detections, with marker and platform composition defined in Section 6.6. The observed map-frame discrepancies were $dx=+0.208$ m (sample SD 0.042 m; range +0.137 to +0.276 m), $dy=+0.030$ m (SD 0.060 m; range −0.049 to +0.127 m), and $yaw=-2.30^\circ$ (SD 4.26°; range −8.99° to +6.99°). These are descriptive summaries of the retained detections, not estimates of map-wide uncertainty. The repeated-marker dependence, opportunistic sampling and post-hoc survey-consistency exclusion limit interpretation. No causal link to a navigation miss, localization improvement or retrospective ground truth for Sessions A–C is established.

### 7.6. Controlled L3b model-substitution replay

The controlled replay tested whether alternative VLMs could operate through the frozen L3b application interface without model-specific changes to its multimodal prompt or production parser. The primary analysis contains 68 E0-excluded, image-complete records, each replayed at three seed bases with the Session E control model and two substitutes, giving 204 outcomes per model. All three models could exercise the interface, but their behavior differed. Parser compliance was 181/204 (88.7%) for qwen3.5:4b, 204/204 (100%) for gemma4:e2b, and 199/204 (97.5%) for qwen3.5:9b. Accepted-node outputs occurred in 160/204, 194/204, and 165/204 outcomes respectively, while archived-node reference agreement was 67.2%, 64.2%, and 58.8%. Same-node agreement across all three seeds occurred for 40/68, 66/68, and 50/68 records. Across the 204 record × seed cells, all three models returned the same node identifier in 97 cases (47.5%). Among the 177 cells with parser-compliant outputs from all three models, agreement on the returned identifier, including the clarification identifier −1, was 97/177 (54.8%). This cross-model disagreement further illustrates why application-interface portability should not be interpreted as semantic model invariance. These are replay-to-archive and within-model stability measures, not semantic-accuracy estimates.

Latency was likewise model dependent on the common replay workstation. Median wall time was 7.20 s for the control, 6.77 s for gemma4:e2b (median ratio 0.94 × to control), and 14.63 s for qwen3.5:9b (2.01 ×). Contract failures numbered 23, 0, and 5 respectively; no transport failure and no retried call occurred in the full acquisition of 621 replay outcomes. On the 13 replayable negation/exclusion records, 117 model-by-seed outcomes were produced and none selected a forbidden node.

The replay also confirms that low-sample memory promotion can remain model/sampling dependent. wet_gloves promoted in all 9 model-by-seed arms, whereas the archived quiet_corner rejection flipped in 8/9 arms; deliveries remained rejected in all nine arms, although that comparison is only a partial multimodal counterfactual because one original image is unavailable. These results support portability of the frozen L3b application interface across the evaluated models. They do not establish semantic model invariance. The cohort is outcome-conditioned on records for which the original system produced an accepted destination and retained the image, so the replay does not estimate alternative-model behavior over all 104 E0-excluded Session E VLM-path attempts.

### 7.7. M3 attribution, selective invocation and decision latency

Both conditions completed their fixed schedules: 342 observations in the standardized-context workload and 252 in the historical-context workload, totaling 594 observations. Outcome distributions and primary item correctness are reported by condition and stratum in Table 6. Counts in the four final-outcome columns are item-seed observations; they are not independent item counts.

On the five familiar items, A produced 15/15 acceptable observations versus 11/15 for B/C, yielding 5/5 versus 4/5 majority-correct items. Two items improved in correct-seed count and three tied: changes from 0/3 to 3/3 and from 2/3 to 3/3 improved only one majority classification. The four wrong executable B/C observations affected two items. VLM calls were 0/15, 12/15 and 15/15 for A, B and C. This descriptive M3 benefit is confined to five familiar items scored against stored associations; no significance or generalization claim follows.

M3 added no semantic or invocation benefit on unseen or clarification-required items. Standardized-context correctness was 25/38 for A and 24/38 for B/C, differing only on familiar associations. B/C decisions matched in all 114 paired observations despite 99 versus 114 VLM calls (69/81 versus 81/81 on unseen formulations): the remaining cascade reduced invocation without observed quality superiority. Failures remained in all denominators: eight observations across six items per arm in the standardized condition, and six across five items in the historical condition. Every clarification was an explicit node_id=−1 response. Clarification-required correctness was 3/6 and 2/5, respectively, without arm differences.

No fast-path decision was incorrect in the standardized condition: 0/27 for A and 0/15 for B. C and all historical-condition arms had no fast-path decisions, so their conditional fast-path error proportion was not applicable. These counts describe the present standardized workload and do not replace or revise the historical E7 G-3 false-fast result.

**Table 6**
Semantic outcomes and VLM invocation in the controlled offline comparison. A, complete system (M3 Step 0, deterministic Steps 1–6, VLM on escalation); B, hybrid system with M3 Step 0 disabled and M1/M2 context retained; C, direct VLM target resolution. All arms share the item context, safety envelope, prompt, response contract and post-response validation. Each item was run with three paired seeds; an item is correct when at least two of its three observations are correct. For familiar items, correctness denotes agreement with the stored historical association; the reference associations were not independently adjudicated. Outcome and VLM-call columns count item-seed observations, which are not independent items. Model/parser failures are retained in every denominator and are never counted as clarification.

| Part A - Standardized context condition (38 items) | | | | | | | |
|---|---|---|---|---|---|---|---|
| Stratum (items) | Arm | Correct items | Acceptable | Incorrect executable | Clarification | Model/parser failure | VLM calls |
| Familiar association (5) | A | 5/5 | 15 | 0 | 0 | 0 | 0/15 |
| | B | 4/5 | 11 | 4 | 0 | 0 | 12/15 |
| | C | 4/5 | 11 | 4 | 0 | 0 | 15/15 |
| Unseen formulation (27) | A | 17/27 | 47 | 22 | 7 | 5 | 69/81 |
| | B | 17/27 | 47 | 22 | 7 | 5 | 69/81 |
| | C | 17/27 | 47 | 22 | 7 | 5 | 81/81 |
| Clarification required (6) | A | 3/6 | 0 | 7 | 8 | 3 | 18/18 |
| | B | 3/6 | 0 | 7 | 8 | 3 | 18/18 |
| | C | 3/6 | 0 | 7 | 8 | 3 | 18/18 |
| All items (38) | A | 25/38 | 62 | 29 | 15 | 8 | 87/114 |
| | B | 24/38 | 58 | 33 | 15 | 8 | 99/114 |
| | C | 24/38 | 58 | 33 | 15 | 8 | 114/114 |
| **Part B - Historical context condition (28 items)** | | | | | | | |
| Stratum (items) | Arm | Correct items | Acceptable | Incorrect executable | Clarification | Model/parser failure | VLM calls |
| Unseen formulation (23) | A | 14/23 | 44 | 17 | 5 | 3 | 69/69 |
| | B | 14/23 | 44 | 17 | 5 | 3 | 69/69 |
| | C | 14/23 | 44 | 17 | 5 | 3 | 69/69 |
| Clarification required (5) | A | 2/5 | 0 | 6 | 6 | 3 | 15/15 |
| | B | 2/5 | 0 | 6 | 6 | 3 | 15/15 |
| | C | 2/5 | 0 | 6 | 6 | 3 | 15/15 |
| All items (28) | A | 16/28 | 44 | 23 | 11 | 6 | 84/84 |
| | B | 16/28 | 44 | 23 | 11 | 6 | 84/84 |
| | C | 16/28 | 44 | 23 | 11 | 6 | 84/84 |

The historical condition contains no familiar-association items. Every clarification was an explicit no-target response from the model.

Table 7 separates decision latency from resolver computation, which is inapplicable to C. Arm summaries are medians of per-item three-seed medians; paired contrasts are computed item by item. Familiar-item A−B latency was −6948.323 ms (95% interval −7076.323 to −0.140 ms); one item also resolved deterministically in B, widening the ratio interval. Across all 38 standardized items, A−B was −17.974 ms (−43.715 to +0.026 ms), providing no overall directional M3 latency advantage. A−C was −29.149 ms (−65.059 to −12.618 ms), with C/A = 1.0042 (1.0017–1.0094). Large savings were confined to inference-avoiding observations; the roughly 52,000–53,000-fold familiar-stratum ratios are not an overall speedup. The standardized B−C interval and all unseen/clarification-required contrast intervals included zero.

In historical context, all arms invoked the VLM for all 84 observations and had identical outcomes. B was modestly faster than A: A−B=+36.904 ms (95% interval +6.924 to +71.570 ms), B/A = 0.9949 (0.9914–0.9989), with the same direction on unseen formulations (Table 7). Routing cannot explain this difference; its backend/runtime cause was not isolated. Historical A−C and B−C intervals included zero. Complete stage-level tables, retaining all outcomes, are provided in the released evidence package (ras-revision-rc1.2).

On the 28 common items, each arm had 16/28 correct in both contexts, but three items improved and three deteriorated in historical context; nodes changed in 27/84 paired observations. Equal aggregate correctness does not establish context robustness. Conditions ran standardized then historical. No thermal slowdown was recorded; GPU power-cap flags occurred in 172/925 and 155/733 samples, respectively, without timing exclusions. Conclusions concern the fixed offline workloads and measured runtime state, excluding live acquisition, navigation and environmental generalization.

Under an independent evaluator's destination judgments for the E7 items, correct items were 14/25 rather than 15/25 in the standardized condition and 10/20 rather than 12/20 in the historical condition, identically in every arm; within the E7 subset, paired arm comparisons remained 0 better, 0 worse and all tied. The independent reference changed absolute correctness, not the comparison between configurations.

The comparison used a uniform typed response contract across arms. A compatibility replay of all 552 saved VLM responses through the frozen deployed parser produced the same accepted integer identifiers and parsing failures for this dataset. This establishes observed-response compatibility, not parser implementation identity or agreement for unobserved response types.

### *7.8. Matcher decision boundaries, exact-match baseline and guard rejection*

At the operational Jaccard threshold of 0.75 and margin of 0.10, the matcher produced 14 correct acceptances, 11 wrong acceptances, nine abstentions when a destination was expected and 13 correct abstentions across the 47 diagnostic items (Table 8). The exact-match baseline accepted none, abstaining on all 23 destination-labeled and 24 abstention-labeled items. This behavior follows from the exclusion of

**Table 7**
Offline decision latency and resolver computation. Part A reports the median over items of their three-seed medians; Parts B/C report item-paired contrasts with stratified paired item-bootstrap 95% intervals (10,000 resamples). Differences are first-arm minus second-arm latencies; negative values indicate lower first-arm latency. For named latency ratios (e.g., B/A), values below 1 indicate lower numerator-arm latency; values above 1 indicate lower denominator-arm latency. All outcomes, including failures, are retained. Live acquisition, robot communication and navigation are excluded; resolver computation is inapplicable to C.

Part A - Arm summaries (median of per-item medians, ms)

| Condition | Stratum (items) | Decision A | Decision B | Decision C | Resolver A | Resolver B |
|---|---|---|---|---|---|---|
| Standardized context | Familiar association (5) | 0.133 | 6948.456 | 7054.761 | 0.116 | 0.249 |
| | Unseen formulation (27) | 6996.319 | 7046.374 | 7045.735 | 0.275 | 0.238 |
| | Clarification required (6) | 9447.636 | 9408.325 | 9500.059 | 0.053 | 0.053 |
| | All items (38) | 6997.017 | 7068.484 | 7052.699 | 0.272 | 0.238 |
| Historical context | Unseen formulation (23) | 7156.539 | 7061.836 | 7126.373 | 0.292 | 0.249 |
| | Clarification required (5) | 7884.172 | 7873.498 | 7957.900 | 0.052 | 0.052 |
| | All items (28) | 7188.215 | 7163.192 | 7155.959 | 0.289 | 0.248 |

Part B - Paired decision-latency contrasts, standardized context condition

| Stratum (items) | Contrast; ratio | Median paired difference, ms | 95% interval | Median paired ratio | 95% interval |
|---|---|---|---|---|---|
| Familiar association (5) | A−B; B/A | −6948.323 | −7076.323 to −0.140 | 52,218.9 | 1.9692 to 56,321.3 |
| | B−C; C/B | −72.576 | −6774.843 to 22.362 | 1.0103 | 0.9966 to 23,754.7 |
| | A−C; C/A | −7054.627 | −7132.946 to −6556.408 | 53,017.8 | 46,778.6 to 56,129.8 |
| Unseen formulation (27) | A−B; B/A | −10.953 | −41.881 to 13.557 | 1.0026 | 0.9942 to 1.0064 |
| | B−C; C/B | −2.902 | −54.883 to 43.290 | 1.0004 | 0.9938 to 1.0076 |
| | A−C; C/A | −26.716 | −62.952 to 16.086 | 1.0036 | 0.9984 to 1.0092 |
| Clarification required (6) | A−B; B/A | −11.885 | −137.318 to 55.517 | 1.0016 | 0.9942 to 1.0154 |
| | B−C; C/B | +39.866 | −98.318 to 166.969 | 0.9946 | 0.9797 to 1.0109 |
| | A−C; C/A | +16.421 | −77.208 to 75.617 | 0.9986 | 0.9892 to 1.0083 |
| All items (38) | A−B; B/A | −17.974 | −43.715 to 0.026 | 1.0041 | 0.9981 to 1.0078 |
| | B−C; C/B | −1.395 | −46.399 to 32.042 | 1.0002 | 0.9955 to 1.0062 |
| | A−C; C/A | −29.149 | −65.059 to −12.618 | 1.0042 | 1.0017 to 1.0094 |

Part C - Paired decision-latency contrasts, historical context condition

| Stratum (items) | Contrast; ratio | Median paired difference, ms | 95% interval | Median paired ratio | 95% interval |
|---|---|---|---|---|---|
| Unseen formulation (23) | A−B; B/A | +51.354 | 13.141 to 94.703 | 0.9926 | 0.9868 to 0.9986 |
| | B−C; C/B | −27.648 | −63.190 to 5.550 | 1.0039 | 0.9993 to 1.0068 |
| | A−C; C/A | +21.541 | −19.492 to 52.390 | 0.9970 | 0.9924 to 1.0026 |
| Clarification required (5) | A−B; B/A | −32.678 | −97.900 to 68.092 | 1.0031 | 0.9907 to 1.0136 |
| | B−C; C/B | +2.401 | −134.084 to 88.040 | 0.9997 | 0.9916 to 1.0184 |
| | A−C; C/A | −44.149 | −231.984 to 82.583 | 1.0049 | 0.9887 to 1.0322 |
| All items (28) | A−B; B/A | +36.904 | 6.924 to 71.570 | 0.9949 | 0.9914 to 0.9989 |
| | B−C; C/B | −17.509 | −54.672 to 10.020 | 1.0024 | 0.9989 to 1.0072 |
| | A−C; C/A | +16.775 | −44.149 to 53.876 | 0.9976 | 0.9936 to 1.0059 |

Paired differences and ratios are medians over items of item-level contrasts; they are not differences or ratios of the Part A arm summaries. Ratios above 100 arise only where one arm resolved deterministically without a VLM call and are not an overall architecture speedup.

**Table 8**
M3 matcher outcomes on the labeled offline diagnostic set of 47 instructions, by item family. The set was deliberately constructed near the decision boundaries and does not estimate performance on a natural instruction distribution. The competing-destination family uses the historical index plus eight declared synthetic association pairs. The exact-match baseline accepts only a unique destination for an identical normalized instruction; normalized exact matches were excluded from the set. Abstention counts items labeled with a destination; correct abstention counts items labeled for abstention.

| Item family (items) | Matcher | Correct acceptance | Wrong acceptance | Abstention, destination expected | Correct abstention |
|---|---|---|---|---|---|
| Valid paraphrase (16) | Jaccard 0.75, margin 0.10 (operational) | 10 | 0 | 6 | 0 |
| | Exact match | 0 | 0 | 16 | 0 |
| | Jaccard 0.60, margin 0.10 (historical reference) | 14 | 0 | 2 | 0 |
| High overlap, different meaning (15) | Jaccard 0.75, margin 0.10 (operational) | 0 | 9 | 2 | 4 |
| | Exact match | 0 | 0 | 2 | 13 |
| | Jaccard 0.60, margin 0.10 (historical reference) | 0 | 13 | 1 | 1 |
| Competing destinations (16) | Jaccard 0.75, margin 0.10 (operational) | 4 | 2 | 1 | 9 |
| | Exact match | 0 | 0 | 5 | 11 |
| | Jaccard 0.60, margin 0.10 (historical reference) | 4 | 2 | 1 | 9 |
| All items (47) | Jaccard 0.75, margin 0.10 (operational) | 14 | 11 | 9 | 13 |
| | Exact match | 0 | 0 | 23 | 24 |
| | Jaccard 0.60, margin 0.10 (historical reference) | 18 | 15 | 4 | 10 |

The 0.60 threshold is reported as a historical reference and was not adopted as an operating setting.

normalized exact matches: the baseline tests replay of stored strings and is not expected to recover paraphrases.

Approximate matching correctly accepted 10/16 valid paraphrases, extending reuse beyond exact replay, but wrongly accepted 9/15 high-overlap, different-meaning items. Table 8 gives the complete family-specific distributions. These deliberately selected cases expose both recovery and lexical-confusion boundaries; they do not estimate performance on a natural instruction distribution.

The historical reference threshold of 0.60 produced 18 correct acceptances, 15 wrong acceptances, four abstentions when a destination was expected and ten correct abstentions. Its four additional correct acceptances were accompanied by four additional wrong acceptances; it was not adopted as a revised operating threshold. All four cases with an exact runner-up gap of 1/10 yielded a floating-point gap just below the 0.10 margin and therefore abstained. This numerical boundary behavior was reported without modifying the frozen matcher. Two of these cases involve the flagged synthetic destination association and should not be interpreted as clean tests of semantic destination validity.

In the deployed-navigator branch test, all 27 forbidden integer proposals across the 14 instructions were rejected, recorded with the negation-violation flag and no navigation dispatch. All 14 permitted-node controls passed the guard and reached intercepted dispatch. The recomputed forbidden envelope matched its transcription for all 14 instructions, and all 139 cases followed the branch anticipated from the frozen code. These counts measure branch conformance, including the weaknesses below, rather than semantic correctness or safety on every case.

The test exposed a parser weakness: a JSON boolean true supplied as node_id reached intercepted dispatch to node 1 in all 14 cases because the deployed parser lacks a type check and Python treats True as equal to 1. No such value was observed among the 91 recorded Session E VLM proposals, all of which were integers. An explicit $-1$ triggered clarification, whereas a missing target was logged as clarification_requested with the distinct missing-answer flag; the latter is a contract failure, not deliberate semantic clarification. String and nonexistent node identifiers were rejected without dispatch. The results establish rejection of the supplied forbidden proposals under the computed envelopes; they do not establish that the VLM avoids forbidden destinations, that arbitrary human negations are interpreted correctly, or that the complete system is safe.

### 7.9. *Development-stage observations and reconciliation of Sessions A–C*

Sessions A–C are retained as development history rather than used to estimate performance of the frozen v4.8-review system. They contain 85, 76, and 4 decision records respectively. Session A is itself version-mixed: 63/85 decisions were recorded under v4.7.4 and 22/85 under v4.8, with the 37 scenario-classified records splitting 26/11 across those versions; Sessions B and C carry the v4.8 development label. Session A also did not run under a uniform loaded-memory state: nine starts recorded three M3 associations and one recorded none; the six-association digest discussed below is the post-Session-A state subsequently loaded in Sessions B and C. None of the 17 A–C audit sessions records the commit-level and clean-working-tree provenance available for Session E, so the development observations are reported only for the software states that the public audit records support.

Across all 85 Session-A decision records, 75 used the deterministic path ($75/85 = 88.2\%$); this is the historical population whose rate numerically matches the submitted "88%" figure, but it uses a different denominator from the 82-record scenario-classified subset. Re-analysis of the public scenario-classified development set corrects the submitted fast-path count from 72/82 to 74/82 (90.2%): 49 decisions used M3 Step 0, 25 another deterministic step, and 8 L3b. The eight L3b cases comprise seven S3new and one S3old case rather than the submitted statement that all were S3new. Session-A scenario counts are likewise corrected to $S2 = 4$ and $S5 = 12$, with the total remaining 37. These values describe the exercised development workload; they are not projected onto Session E or interpreted as a general fast-path probability.

The development logs also show repeated deterministic reuse of stored instruction–node associations. Session B contains 33 M3 reuse decisions on Xplorer-B across four stored instruction–node mappings: 14 restroom, 10 fire-safety, 6 fresh-air, and 3 sit-and-relax repetitions, drawn from a six-association post-Session-A digest. They are therefore repeated reuse observations over a small stored set, not 33 independent transfer cases and not a basis for a transfer-accuracy confidence interval. The historical digest state identified by the 97241265… content identifier is inspectable and its loading/use in Sessions B/C is auditable, but it cannot be regenerated end to end solely from the 17 public A–C audit logs because it was compiled from a broader development decision history. The revision therefore uses Session E E4/E4b, rather than this historical digest lineage, as the confirmatory evidence for the promotion-and-reuse mechanism.

Other submitted development-stage quantities are corrected under the same policy rather than silently carried forward. The previously quoted mean position error over five missed A–C runs is 3.97 m rather than 3.77 m; the complete deterministic cascade is described as sub-millisecond rather than by the smaller round figure previously quoted; and promotion-frequency/consistency values are reported only with their corresponding population and rounding convention. These corrections do not alter the role assigned to A–C: they preserve transparent development evidence while the frozen Session E campaign provides the confirmatory basis for the final-system claims.

## 8. Discussion and limitations

### 8.1. What the frozen evidence establishes

The evidence supports SAS as a higher-layer integration framework over the inherited L0–L2 navigation/perception foundation. Session E evaluates deterministic-before-VLM resolution, guard and clarification behavior, post-arrival confirmation, M3 promotion and reuse, and physical operation on two robot configurations. The controlled A/B/C comparison complements this evidence by isolating M3 target reuse with M1/M2 context held constant and comparing selective resolution with direct VLM target resolution. M1/M2 are removed jointly with M3 in E6; their individual contributions and those of M4/M5 remain unisolated.

Within those limits, the results establish a coherent operational pattern. Deterministic resolution can return selected targets with sub-millisecond resolver computation when an admissible rule or stored association applies; unresolved, ambiguous, negated or conflicting instructions can be escalated instead of being forced through that path. Controlled present/absent trials show that post-arrival confirmation can remain distinct from physical arrival, and the E4b sequence demonstrates one complete association lifecycle from VLM-mediated induction through promotion, digest compilation and installation to deterministic reuse on the other robot platform. The concurrency block further shows functional two-robot access to the shared inference service, while the controlled replay shows that the frozen L3b application interface can be exercised by alternative VLMs without model-specific prompt or parser changes. These findings validate the tested higher-layer integration and mechanisms under the frozen configuration; they do not constitute complete validation of all SAS layers, environments, memory classes or model behaviors.

### 8.2. Selective reasoning: efficiency versus semantic coverage

The principal benefit of the deterministic-first design is selective computational economy, not universal replacement of model reasoning. In Session E, 112 E0-excluded decision cycles resolved without a model call, but that fraction is a property of the exercised block composition rather than an estimate for naturally occurring indoor instructions. The sealed E7 set makes this distinction explicit: only 5 of 28 unseen instruction cells used the deterministic fast path, and one of those five reached a non-acceptable destination. Conversely, the repeated E3 intents were all resolved deterministically and with an invariant resolver path. Fast-path coverage therefore depends strongly on lexical and semantic proximity to rules, graph entities and compiled associations.

Selective invocation has its largest timing benefit when an admissible deterministic decision avoids VLM inference. The naturally selected Session E paths contain different instructions and do not provide a causal VLM-only baseline. The controlled comparison in Sections 6.8 and 7.7 supplies matched instruction/context contrasts: M3 improves stored-destination correctness on the five familiar items, while B and C make identical decisions in all 114 standardized-context observations despite 99 versus 114 VLM calls. Large latency reductions are concentrated in inference-avoiding paths; they are not a uniform architecture speedup. E6 separately shows the effect of removing the compiled memory layer, with M3 and M1/M2 removed jointly. The evaluated routing boundary remains imperfect: E7 includes incorrect executable decisions and three L3b parsing failures, with no explicit clarification requests. Stronger deterministic discrimination and typed response validation can be developed within L3, but their semantic and timing effects require evaluation beyond v4.8-review.

### 8.3. Memory reuse: stability, correctness, and transfer scope

The memory results separate three questions that are easily conflated: whether an association recurs, whether it is semantically correct, and whether it can be reused across platforms. The promotion rule addresses only the first. It promotes repeated instruction–node outcomes when their frequency and dominant-destination consistency satisfy the frozen gate and at least one accepted L3b resolution is present. The Session E candidates illustrate both sides of that rule: wet_gloves is stable enough to promote, whereas quiet_corner and deliveries are rejected in the archived run. The controlled replay then exposes the low-sample boundary directly: wet_gloves remains promoted in all nine model-by-seed arms, while the archived quiet_corner rejection flips in eight of nine. Promotion must therefore be understood as a sampled-output stability criterion, not as evidence that the promoted destination represents semantic truth or a user-confirmed preference.

That distinction is also the main safety limitation of deterministic reuse. Once promoted, an association can bypass the VLM at Step 0, so a stable but semantically wrong association could become a fast deterministic rule. The frozen resolver mitigates one important failure mode by giving negation and exclusion precedence over Step-0 acceptance and by validating forbidden destinations after L3b, but these safeguards do not convert the promotion gate into a correctness oracle. Explicit user feedback or another independent semantic-validity signal would be required to make stronger preference-learning claims.

E4b demonstrates one induced association, deliberately compiled and installed in a shared digest, followed by four deterministic reuse cycles on the recipient platform. These cycles verify one mapping rather than four independent associations or autonomous robot-to-robot learning. E6 shows a resolution change in four paired units after joint removal of M3 Step 0 and the M1/M2 prompt prefix, so its changed L3b destinations cannot be attributed to M3 alone. The controlled A–B comparison in Sections 6.8 and 7.7 provides complementary M3 attribution with M1/M2 held constant, confined to the five familiar stored associations. Independent M1/M2 effects and M4/M5 performance contributions remain unevaluated.

### 8.4. Model portability and shared-inference deployment

The model-substitution replay narrows the portability claim to the level actually exercised by the evidence. Across three evaluated VLMs and three seed bases, the same frozen multimodal prompt, parser and context contract remained usable without model-specific adaptation, but parser compliance, abstention, destination selection, seed unanimity and latency differed by model. Agreement with the archived destination is only reference agreement against one prior stochastic outcome, not semantic accuracy. Moreover, the 68-record primary replay cohort is outcome-conditioned on original accepted-node cases with retained images, so it does not estimate substitute-model behavior over all Session E L3b attempts. The supported conclusion is therefore L3b application-interface portability across the evaluated models, not semantic model invariance or model-agnostic reasoning.

The deployment architecture also places that result in context. Locomotion, sensing and metric navigation execute onboard, whereas L3a deterministic resolution, L3b inference and L5 audit/memory services execute on the shared workstation; a GPU is required for L3b inference, not for the deterministic resolver. The system should consequently be described as a distributed robot–workstation deployment rather than as fully onboard GPU-free semantic inference. E5 shows that two robot configurations can use the shared L3b service concurrently in three tested pairs, with all six missions completing and all responses parsing, but latency was higher than in the descriptive non-concurrent comparator. The available instrumentation does not identify whether that increase arises from contention, partial serialization or another shared-resource effect. The evidence therefore demonstrates small-scale concurrent operation, not fleet-scale throughput or a measured server-serialization model.

### 8.5. Limitations and threats to validity

Several limitations bound the interpretation of the revised results. First, external validity is narrow. The confirmatory campaign uses two robot configurations in one indoor environment and one route graph; the sealed E7 evaluation broadens linguistic challenge but runs only on Xplorer-B and does not test a second building, unseen map or substantially different environment. In particular, the campaign did not vary POI density, test symmetric spatial layouts in which Step-6 proximity may remain ambiguous, or systematically impose dynamic human/object occlusion. The work therefore characterizes semantic-resolution behavior and higher-layer integration at the tested site, without establishing environmental generalization or fleet-scale deployment. Similarly, no benchmark-equivalent head-to-head experiment with the IROS system, Hydra-Nav or other published systems was performed, and the naturally escalated Session E L3b cases are conditioned by the SAS cascade. The all-instruction VLM-only target-resolution baseline in Sections 6.8 and 7.7 applies to the fixed offline workload and does not provide a physical-navigation baseline. Cross-paper comparisons are consequently architectural and qualitative rather than performance rankings.

Second, construct and causal validity remain limited in several subsystems. The M3 gate measures recurrence of model-derived destinations rather than user-confirmed semantic preference; the frozen experiment contains no operator-confirmation field or user-ground-truth preference mechanism, so stable-but-wrong promotion remains possible. E6 removes M3 Step 0 together with the M1/M2 prompt prefix, preventing independent attribution of destination changes within that block; the controlled comparison in Sections 6.8 and 7.7 addresses M3 target reuse with M1/M2 context held constant; the cascade analysis reports which step resolved each instruction first rather than the causal effect of removing a step; and M4/M5 are not independently evaluated. Isolated M1/M2 context interventions, causal single-rule removals within the L3a cascade, removal of static-object prioritization and a full visual-input on/off ablation remain untested. The eight E1b trials provide a controlled fire-extinguisher presence/removal result; they do not test a present wall-mounted object outside the camera field of view. Broader confirmation counts remain descriptive because observability and semantic ground truth are not uniformly controlled.

The controlled comparison isolates the contribution of M3 target reuse while retaining M1/M2 context and common validation safeguards. For the five familiar items, M3 reuse increased agreement with the stored associations. These reference associations were not independently adjudicated. The comparison does not establish independently learned human preferences or broad M3 generalization to new formulations, and no arm improved clarification performance on the tested cases. Archive completeness limits the workload: nine items lacking a start image were excluded without substitution, and missingness is not assumed random. The historical condition covers only 28 items and contains no familiar-association stratum. Offline decision latency includes archived-image preparation when needed, but excludes live acquisition, robot communication and motion, resident-state preparation and model startup/preload. The runner-isolated, resident-model operating state is distinct from production cache throughput or cold-start latency. The conditions were run standardized then historical; observed power-cap flags and small timing differences remain part of the reported runtime state. Equal decisions or aggregate correctness do not establish equivalence or robustness, and no physical-navigation benefit follows from this experiment.

The matcher-boundary evaluation is a deliberately constructed diagnostic set, with access to similarity scores during construction; it does not estimate general paraphrase accuracy or establish a universally appropriate threshold. Its synthetic competing-association index and two flagged destination associations further restrict semantic interpretation. The guard test verifies rejection of supplied forbidden proposals under the computed envelopes, without validating arbitrary negation interpretation or complete system safety. The deployed parser does not type-check the node identifier: boolean true reached intercepted dispatch as node 1 in 14/14 synthetic cases, although all 91 recorded Session E VLM proposals were integers. The frozen deployed implementation was not modified; branch conformance therefore includes this weakness.

Third, physical outcomes do not isolate platform effects because the hardware differences in Table 3 were not experimentally controlled. The AprilTag reference is a contemporaneous Session E check based on 11 detections, ten from one marker, with a post-hoc survey-consistency exclusion. Counts, means, sample standard deviations and observed ranges characterize map-frame discrepancies; repeated-marker detections are not independent. This reference neither identifies a navigation-miss cause nor retrospectively ground-truths Sessions A–C. No localization improvement was implemented or evaluated. The discrepancy motivates frame-alignment and camera/robot-extrinsic checks and broader independent reference coverage before attributing an error to AMCL.

Reproducibility differs across the evidence history. Session E has a frozen commit, clean-tree record and fixed model/configuration identity; Sessions A–C lack commit-level provenance and remain development-stage evidence. The historical post-Session-A digest is inspectable and its use in B/C auditable, but cannot be regenerated end to end from the 17 public A–C logs alone. The public release distinguishes rerunnable analyses from inspectable artifacts and executions requiring protected modules, as detailed in the Data Availability Statement and repository reproducibility notes. In particular, public post-run analysis does not make semantic-comparison or guard execution independently rerunnable: these require protected implementations, and the comparison also requires the station runtime. Camera statistics are recomputable; reacquisition requires the robot and its service. Offline analyses and replays complement, rather than extend, the physical-navigation campaign.

## 9. Conclusions

SAS provides a six-layer integration framework connecting semantic interpretation, mission execution and operational experience with a ROS 2/Nav2 navigation and perception foundation. The contribution developed here lies principally in L3–L5: a deterministic–VLM reasoning cascade, mission orchestration and an operational-memory lifecycle that makes recurrent instruction–destination associations available for later deterministic reuse. Explicit boundaries between semantic selection, validation, physical navigation and post-arrival confirmation allow these mechanisms to be evaluated and developed within a common architecture.

The frozen v4.8-review implementation demonstrates this integration on two physical robot configurations in one indoor environment. In Session E, 112 of 216 decision cycles outside the start gate resolved without a VLM call, with sub-millisecond resolver-compute latency. The sealed instruction challenge yielded 17/28 acceptable outcomes; its five fast-path decisions included one incorrect selection. A promoted association was compiled and deliberately installed for reuse on the recipient platform. The controlled offline comparison adds direct attribution evidence: on five familiar items, the full system agreed with the stored associations on 5/5 items versus 4/5 for the M3-disabled hybrid and direct-VLM configurations and avoided all 15 familiar-item VLM calls. In standardized context, the cascade without M3 reduced VLM invocation while producing the same decisions as direct VLM resolution. Large latency reductions were concentrated in familiar paths that avoided inference; they do not describe a uniform speedup across instructions.

The diagnostics identify different kinds of boundaries. Partial object-class matching, lexical-overlap acceptance errors and response-type handling are limitations of the evaluated L3 rules and parser. The M3 promotion gate has a separate methodological boundary: repeated consistent outputs can still encode a semantically wrong association. SAS supplies defined integration points for typed response validation, stronger semantic discrimination and independent confirmation before association promotion. These are concrete development routes within L3 and L5; their effect on decision quality and latency requires evaluation in subsequent versions. The present guard tests establish behavior for the supplied proposals and computed exclusion envelopes, while the physical outcomes distinguish destination correctness, arrival and confirmation.

The evidence therefore supports SAS as an implemented framework for selective hybrid semantic reasoning and reusable operational memory, with benefits tied to the mechanisms and workloads that were tested. The two-platform campaign establishes physical integration and managed shared-memory reuse; broader environmental and instruction coverage, separate evaluation of the remaining memory components and fleet-scale exchange require additional studies. The model-substitution replay establishes observed compatibility of the L3b application interface without establishing semantic equivalence across models. Publicly rerunnable analyses and inspectable frozen artifacts support scrutiny of these findings, with protected-code dependencies stated separately from the architectural and experimental scope.

## Declaration of generative AI and AI-assisted technologies in the manuscript preparation process

During the preparation of this work the authors used Anthropic Claude in order to support structural organization, literature search assistance, and editorial refinement. After using this tool, the authors reviewed and edited the content as needed and take full responsibility for the content of the publication.

## Funding

This research was supported by the National University of Science and Technology POLITEHNICA Bucharest (UNSTPB).

## CRediT authorship contribution statement

**Bogdan Felician Abaza:** Writing – review & editing, Writing – original draft, Visualization, Validation, Supervision, Software, Resources, Project administration, Methodology, Investigation, Formal analysis, Data curation, Conceptualization. **Andrei-Alexandru Staicu:** Writing – review & editing, Visualization, Software, Investigation, Data curation, Validation. **Cristian Vasile Doicin:** Writing – review & editing, Visualization, Validation, Supervision, Resources, Funding acquisition, Formal analysis.

## Declaration of competing interest

The authors declare that they have no known competing financial interests or personal relationships that could have appeared to influence the work reported in this paper.

## Data availability

The evidence and reproducibility package supporting this work is publicly available at https://github.com/bogdan-abaza/nav2-sas-vlm-memory, with the revised package identified by the frozen release tag ras-revision-rc1.2. It retains the development-stage Sessions A–C records and the frozen Session E campaign evidence and historical analysis artifacts, and adds the corrected E7 taxonomy and navigation denominators, controlled semantic-comparison records, labeled matcher-boundary set, guard branch-test records, independent E7 adjudication, descriptive AprilTag summary and separate camera-transaction measurement. Original frozen datasets and analytical artifacts are preserved, and every added component has an integrity manifest.

The repository's reproducibility notes distinguish analyses rerunnable from the released files, frozen artifacts that can be inspected and verified, and protected implementations. Publicly rerunnable analyses include the Session E core and corrective reanalysis, M3 matching-parameter and boundary analyses, independent-adjudication analysis, controlled-comparison post-run analysis and reference sensitivity, and AprilTag summary. Camera summary statistics are recomputable from the released records. The historical cascade-step and promotion-threshold supporting packages can be inspected and verified, but their complete original reproduction workflows cannot be run from the release alone. Repeating the controlled semantic-comparison execution, guard branch test or memory extraction requires protected modules; repeating camera acquisition requires the robot and its service. The model-substitution replay package separately documents its public analysis and runtime requirements. The complete navigator, executive contract, memory extractor and associated protected modules remain available to the editor and reviewers in confidence upon request through the journal.